 \documentclass[final,1p,times,authoryear]{elsarticle}

\usepackage{amssymb}
\usepackage{amsmath}
\usepackage{subcaption}
\usepackage{amsmath}
\usepackage{amssymb}  
\usepackage{makecell}
\usepackage{pifont}
\usepackage{enumitem}
\usepackage{hyperref}
\usepackage{caption}
\usepackage{adjustbox}
\usepackage{float}
\usepackage{multirow}
\usepackage{comment}
\usepackage{collcell}
\usepackage{enumitem}
\usepackage{wasysym}
\usepackage{xcolor}
\usepackage{booktabs, pifont}
\newcommand{\cmark}{\textcolor{green!60!black}{\ding{51}}}%
\newcommand{\xmark}{\textcolor{red!70!black}{\ding{55}}}%
\usepackage{fontawesome5}
\usepackage{soul}    
\usepackage{xcolor}  
\usepackage{arydshln}
\usepackage[table,xcdraw]{xcolor}

\newif\ifhighlight
\highlightfalse 

\newcommand{\myblue}[1]{%
  \ifhighlight
    \myblue{#1}
  \else
    #1
  \fi
}

\newcommand{\BlueIfHighlight}[1]{%
  \ifhighlight
    \textcolor{blue}{#1}%
  \else
    #1%
  \fi
}

\newcolumntype{b}{>{\collectcell\BlueIfHighlight}c<{\endcollectcell}}

\begin{document}

\begin{frontmatter}



 \cortext[cor1]{Corresponding author}

\title{Explicitly Aligned Multi-View Breast Cancer Risk Prediction from Variable-Length Screening Histories}


\author[a]{Solveig Thrun\corref{cor1}} \ead{solveig.thrun@uit.no}    
\author[a]{Zijun Sun}
\author[a]{Suaiba A. Salahuddin} 
\author[a]{Kristoffer Wickstrøm} 
\author[a]{Elisabeth Wetzer} 
\author[b]{Stine Hansen} 
\author[a,c,d]{Robert Jenssen} 
\author[a,c]{Michael Kampffmeyer} 

\affiliation[a]{organization={Department of Physics and Technology, UiT The Arctic University of Norway},
            city={Tromsø},
            postcode={9037}, 
            country={Norway}}
\affiliation[b]{organization={SPKI The Norwegian Centre
for Clinical Artificial Intelligence, University Hospital of North Norway},
            city={Tromsø},
            postcode={9019}, 
            country={Norway}}    

\affiliation[c]{organization={Norwegian Computing Center},
            city={Oslo},
            postcode={0373}, 
            country={Norway}}
\affiliation[d]{organization={Pioneer Centre for AI, University of Copenhagen},
            city={Copenhagen},
            postcode={1350}, 
            country={Denmark}}     
            
\begin{abstract}
Accurate breast cancer risk prediction from screening mammography is essential for improved risk stratification and personalized screening. Such prediction can benefit from both temporal changes in breast tissue captured across longitudinal examinations and complementary anatomical information provided by craniocaudal (CC) and mediolateral oblique (MLO) views. Effectively leveraging longitudinal information requires identifying corresponding tissue regions across examinations, yet most existing approaches rely on implicit feature alignment without explicitly enforcing spatial consistency. Recent work has demonstrated the benefits of explicit temporal alignment; however, existing explicit alignment methods are not leveraging multi-view information and instead model each view independently, preventing them from jointly exploiting the complementary information provided by CC and MLO views. Further, explicit approaches have been limited to only leveraging one prior examination. To fill these gaps, we propose
\href{https://github.com/sot176/FlexLMV-Net}{FlexLMV-Net} \footnote{\url{github.com/sot176/FlexLMV-Net}}, a breast-wise longitudinal multi-view framework that jointly models CC and MLO views across screening examinations while accommodating a variable number of prior examinations. FlexLMV-Net explicitly aligns longitudinal feature maps and integrates complementary views through cross-view temporal fusion and a dual-stream attention block. Evaluated on EMBED and CSAW-CC, FlexLMV-Net consistently outperforms existing approaches across breast density, cancer subtypes, and demographic subgroups, highlighting the benefits of explicit longitudinal alignment, multi-view modeling, and flexible use of available screening history.
We further introduce \href{https://github.com/sot176/BCRBench}{BCRBench} \footnote{\url{github.com/sot176/BCRBench}}, an open-source benchmark that unifies state-of-the-art single- and multi-timepoint breast cancer risk prediction models under standardized preprocessing, training, inference, and evaluation protocols, enabling fair and reproducible comparison of existing methods.
\end{abstract}

\begin{keyword}
Breast cancer risk prediction \sep Screening mammography \sep Longitudinal analysis \sep Multi-view learning


\end{keyword}

\end{frontmatter}



\section{Introduction}\label{intro}
Breast cancer is the most commonly diagnosed cancer among women worldwide and remains a leading cause of cancer-related mortality~\citep{cancerstat}. Population-based mammography screening programs have contributed substantially to reductions in breast cancer mortality by enabling earlier cancer detection and timely treatment~\citep{zielonke2020evidence}. In routine clinical practice, each screening examination consists of two standard mammographic views per breast, the craniocaudal (CC) and mediolateral oblique (MLO) views, which provide complementary anatomical information for lesion detection~\citep{pharmaceutics13050723}. These examinations are typically performed at regular intervals, with current screening guidelines recommending annual or biennial screening primarily based on age~\citep{ren2022guidelines}. While this population-based approach has proven effective, it also results in millions of negative examinations each year, increasing healthcare costs, unnecessary recalls, overdiagnosis, and patient burden~\citep{habib2021recommendations,bond2013psychological,mccann2002}. Conversely, these screening protocols may result in under-screening of women at high risk of breast cancer, delaying the detection of early malignancies and leading to diagnosis at more advanced stages~\citep{rubio2024risk}. Furthermore, age-based screening intervals do not adequately account for the substantial variation in individual breast cancer risk~\citep{clift2022}.

These limitations have motivated a shift toward personalized screening strategies, in which screening frequency and supplemental imaging are tailored to an individual's estimated risk of developing breast cancer. Such individualized screening strategies have the potential to improve early cancer detection, reduce screening-related harms, and optimize the use of healthcare resources~\citep{clift2022,vanwinkel2025}.

Recent advances in deep learning have enabled future breast cancer risk to be estimated directly from mammograms, often outperforming traditional clinical risk models~\citep{tyrer2004} based on demographic and lifestyle factors. Early deep learning approaches primarily relied on the current screening examination alone~\citep{mirai,asymmirai}, learning imaging biomarkers associated with future cancer development from a single time point. However, radiologists routinely compare the current mammogram with prior examinations to identify subtle tissue changes that may indicate early malignant transformation~\citep{akwo2024access, prioirmammo}. Motivated by this clinical workflow, recent models have begun incorporating prior mammograms, allowing them to model temporal changes in breast tissue and improve long-term risk prediction~\citep{primeplus,Wang2023,lrp_net, Kar_Longitudinal_MICCAI2024,Wan_Ordinal_MICCAI2024,vmramar}.

While most existing longitudinal mammography models compare current and prior mammograms using implicit feature alignment (e.g., LoMaR~\cite{Kar_Longitudinal_MICCAI2024}, VMRA-MaR~\cite{vmramar}), recent work has demonstrated that explicitly aligning mammograms before temporal feature comparison produces more accurate representations of longitudinal tissue changes and improves breast cancer risk prediction~\citep{thrunarxiv2025}. However, existing explicit alignment approaches, including ImgFeatAlign~\citep{mammoregnet} and OA-BreaCR~\citep{Wan_Ordinal_MICCAI2024}, perform longitudinal modeling independently for each mammographic view, thereby not exploiting the complementary information contained in the CC and MLO views during temporal analysis. Moreover, existing longitudinal approaches typically assume a fixed number of prior examinations, despite substantial variation in the availability of prior screening mammograms across women. This limits their applicability to patients with different screening histories and, in particular, to women attending screening for the first time.

To address these limitations, we propose FlexLMV-Net, a breast-wise longitudinal multi-view model for breast cancer risk prediction that jointly exploits temporal and complementary multi-view information while accommodating a variable number of prior examinations. Our model explicitly aligns feature maps from prior and current mammograms for both CC and MLO views, enabling accurate modeling of longitudinal tissue changes, and integrates the aligned representations through cross-view temporal fusion. By combining explicit temporal alignment, joint multi-view analysis, and flexible handling of longitudinal histories, FlexLMV-Net can leverage the available screening history without requiring a fixed number of prior examinations, providing a more comprehensive representation of breast tissue evolution for breast cancer risk prediction.

The main contributions of this study are summarized as follows:
\begin{itemize}
 \item We propose FlexLMV-Net, a breast-wise longitudinal multi-view framework that captures temporal changes through explicit feature-space alignment and cross-view fusion while accommodating variable numbers of prior examinations.
\item We introduce a multi-view dual-stream attention block that models intra- and inter-view dependencies through parallel self- and cross-attention, enabling temporal fusion across variable-length screening histories.
\item We validate our approach on two large-scale public datasets (EMBED and CSAW-CC), demonstrating robust performance across breast density, cancer subtypes, and demographic subgroups.
\item We contribute BCRBench\footnote{\url{github.com/sot176/BCRBench}}, an open-source benchmark  that unifies state-of-the-art models for single- and multi-timepoint breast cancer risk prediction within a common, reproducible framework, supported by comprehensive documentation\footnote{\url{https://sot176.github.io/BCRBench}}, to enable fair and reproducible model comparison.

\end{itemize}

A preliminary version of this work appeared in~\citep{ownpaperMICCAI2026}. 
In this extended study, we substantially enhance our previous model (LMV-Net) 
through the following key improvements:
\begin{enumerate}
    \item We extend LMV-Net to FlexLMV-Net, enabling variable numbers of prior examinations through a feature-aware temporal fusion block that integrates available longitudinal information. This allows FlexLMV-Net to also be applied to first-time screening participants, broadening its applicability to the wider screening population.
    \item We introduce BCRBench, a structured benchmark for image-based breast cancer risk prediction that provides standardized evaluation protocols and facilitates reproducible comparison of risk prediction models.
    \item We deepen the literature review to provide a more comprehensive positioning of our approach relative to existing methods and to clarify how our work extends prior research. 
    \item We substantially expand the methodological description of FlexLMV-Net, detailing its handling of variable-length longitudinal inputs, explicit feature alignment, and multi-view temporal fusion, while systematically analyzing the contribution of each component to predictive performance.
    \item We further assess the generalization of LMV-Net and FlexLMV-Net across datasets and evaluate its robustness across clinically relevant subgroups, demonstrating its applicability across different populations and imaging settings.
\end{enumerate}

The paper is organized as follows. Section~\ref{relwork} provides an overview of relevant related work, focusing on breast cancer risk prediction and multi-view analysis. Section~\ref{methods} presents the proposed FlexLMV-Net in detail. In Section~\ref{expsetup}, we describe the experimental setup and the evaluation metrics used to assess model performance. Sections~\ref{result} presents and discusses the results. Finally, Sections~\ref{limitationsoutlook} and~\ref{conclusion} address the limitations of the proposed approach, outline directions for future research, and summarize the key findings of the study.

\section{Related Work}\label{relwork}

\subsection{Breast Cancer Risk Prediction} \label{relworkriskpred}

Traditional breast cancer risk prediction has relied on classical models such as Tyrer--Cuzick~\citep{tyrer2004}, which primarily use manually collected clinical variables and do not leverage imaging information from routine screening. However, more recently end-to-end deep learning approaches have enabled risk prediction directly from full-field mammograms~\citep{mirai,Kar_Longitudinal_MICCAI2024,primeplus,Wan_Ordinal_MICCAI2024,vmramar,trinet2025}, allowing models to learn predictive imaging biomarkers automatically~\citep{2024riskreview}. Mirai~\citep{mirai} established a strong baseline using multi-view mammograms and significantly outperformed traditional risk models for long-term risk prediction. AsymMirai~\citep{asymmirai} further demonstrated that bilateral asymmetry can serve as a predictive and interpretable imaging biomarker. However, these approaches operate on a single screening timepoint and do not capture temporal changes in breast tissue that may provide additional information about future cancer risk.

Motivated by the longitudinal assessment routinely performed by radiologists, several recent methods incorporate prior screening examinations into risk prediction. PRIME+~\citep{primeplus}, VMRA-MaR~\citep{vmramar}, LoMaR~\citep{Kar_Longitudinal_MICCAI2024}, and TRINet~\citep{trinet2025} combine information from current and prior mammograms, demonstrating improved risk stratification compared with single-timepoint approaches. TRINet additionally incorporates time-decay attention and handcrafted radiomic features within a multiple-instance learning framework. However, these methods primarily rely on implicit alignment, comparing or combining representations across timepoints without explicitly enforcing spatial correspondence, which may limit their ability to capture subtle longitudinal tissue changes.

While implicit alignment is computationally efficient, it is sensitive to geometric variations caused by differences in patient positioning and breast compression across visits. Recent evidence indicates that explicit spatial registration can better capture longitudinal tissue changes and improve risk prediction~\citep{thrunarxiv2025}. OA-BreaCR~\citep{Wan_Ordinal_MICCAI2024} learns deformation fields directly in feature space, while ImgFeatAlign~\citep{mammoregnet,thrunarxiv2025} estimates image-based deformation fields and applies them to feature representations, enabling more precise spatial correspondence across examinations.

Despite these advances, existing explicit alignment approaches model mammographic views independently and therefore do not jointly exploit the complementary information provided by CC and MLO projections during longitudinal analysis. Furthermore, existing explicit longitudinal models typically assume a fixed number of prior examinations, limiting their applicability to patients with different screening histories and to examinations where no prior mammogram is available. FlexLMV-Net addresses these limitations by explicitly aligning longitudinal representations, jointly modeling CC and MLO views, and accommodating a variable number of prior examinations.

\subsection{Multi-view mammography analysis} \label{multiview}

Early deep learning approaches in multi-view mammography typically processed CC and MLO projections through independent convolutional branches, combining them only after global pooling or simple feature concatenation~\citep{wu2020dnnradiologists}. While effective, such strategies may discard richer cross-view correlations that radiologists exploit clinically. More recent architectures explicitly model these interactions to capture cross-view relationships~\citep{vantulder2021multiview}. For instance, MV-Swin-T~\citep{mvswinT} introduces a Multi-headed Dynamic Attention Block, enabling self- and cross-view information fusion at the spatial feature level and outperforming multi-view CNN baselines across multiple benchmark datasets. Similarly, MVPT-NET~\citep{mvpnet} leverages visual prompt tuning on a pre-trained single-view backbone to integrate multi-view information efficiently without requiring full retraining.

In longitudinal risk prediction, multi-view information has also been combined with temporal modeling. LRP-NET~\citep{lrp_net} processes CC and MLO projections from multiple consecutive examinations through parallel GRU branches, capturing bilateral spatiotemporal tissue differences for near-term risk estimation. LoMaR~\citep{Kar_Longitudinal_MICCAI2024} extends the single-timepoint Mirai model with a transformer-based visit aggregator and temporal positional encodings, distilling information from multiple visits and all four standard views into a comprehensive risk embedding. VMRA-MaR~\citep{vmramar} further incorporates bilateral asymmetry through side encoding and temporal attention, capturing fine-grained spatial and longitudinal asymmetric cues across CC and MLO views.

Despite these advances, existing multi-view longitudinal models either do not explicitly register mammographic views across timepoints or assume a fixed number of prior examinations. This limits their ability to jointly exploit spatially aligned multi-view information while accommodating patients with different screening histories. FlexLMV-Net addresses these gaps by combining explicit longitudinal feature alignment, cross-view temporal fusion, and flexible input handling for varying numbers of prior examinations within a unified framework.

\section{Methodology}\label{methods}
\begin{figure}[htbp]
    \centering
    \includegraphics[width=0.6\linewidth]{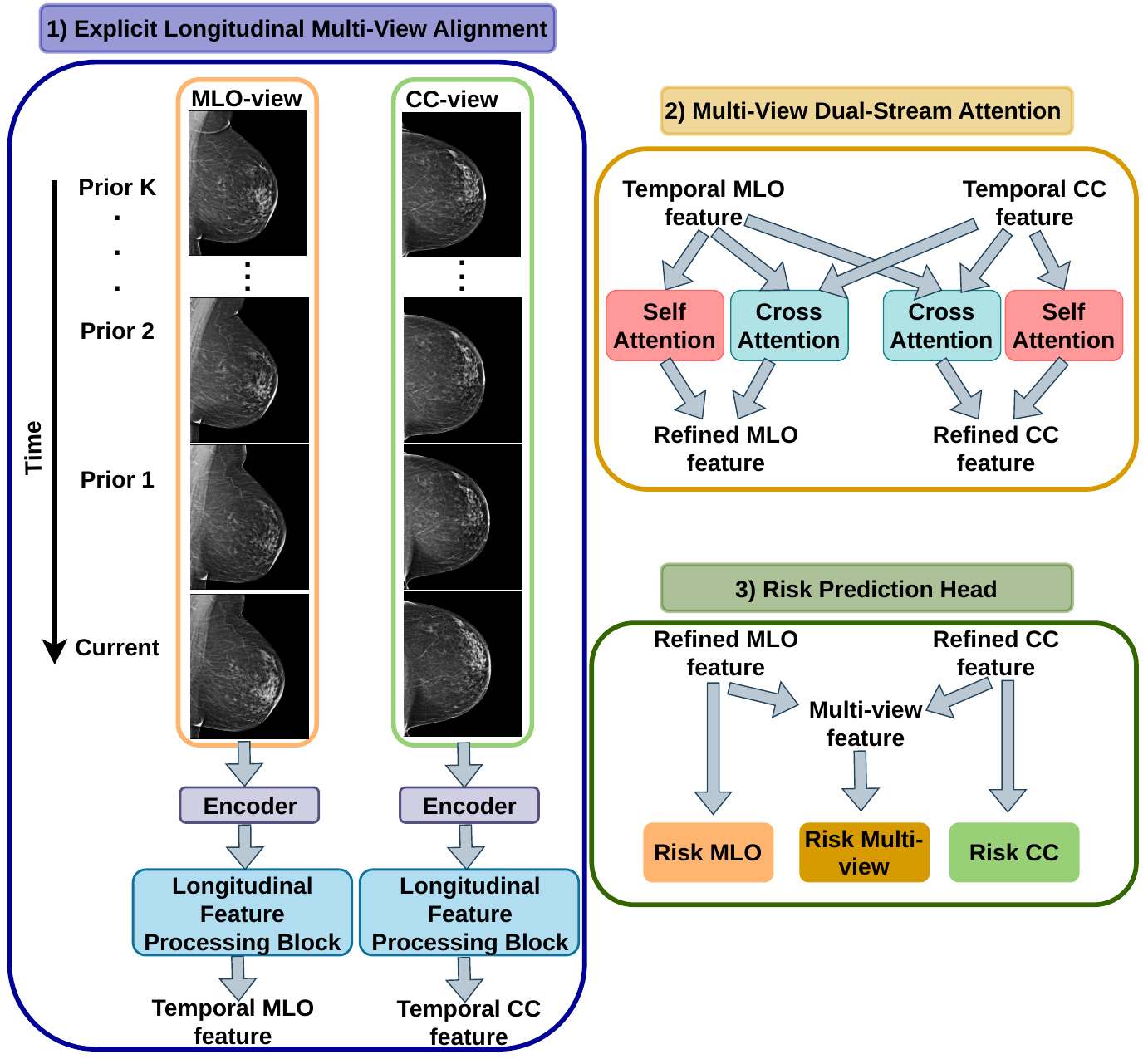}
    \caption{Overview of the proposed FlexLMV-Net architecture. The framework consists of three main components: (1) explicit longitudinal multi-view alignment to establish correspondence between current and prior examinations, (2) a multi-view dual-stream attention block that models intra-view and inter-view dependencies across longitudinal data, and (3) a risk prediction head that leverages fused multi-view representations, with auxiliary view-specific supervision during training, to predict five-year breast cancer risk.}
    \label{fig:lmvnet_highlevel}
\end{figure}

Breast cancer risk prediction estimates the probability of developing breast cancer within a fixed time horizon. Here, we consider five-year risk prediction from screening mammography. Longitudinal data provide valuable information by enabling the detection of subtle tissue changes over time. To jointly exploit longitudinal and multi-view (CC and MLO) information while accommodating variable numbers of prior examinations, we introduce FlexLMV-Net (Figure~\ref{fig:lmvnet_highlevel}). The framework comprises three components. First, longitudinal feature maps from prior examinations are explicitly aligned with the current examination to establish spatial correspondence and capture temporal changes, then aggregated by the Feature-Aware Temporal Fusion block. Second, a multi-view dual-stream attention block models intra- and inter-view dependencies through self- and cross-attention, integrating complementary information across views and longitudinal time points. Finally, a risk prediction head estimates five-year risk from a fused multi-view representation, with view-specific features used as auxiliary supervision during training. Each component is described in detail in the following subsections.

\subsection{Explicit Longitudinal Multi-View Alignment}

\begin{figure}[htbp]
    \centering
    \includegraphics[width=0.99\linewidth]{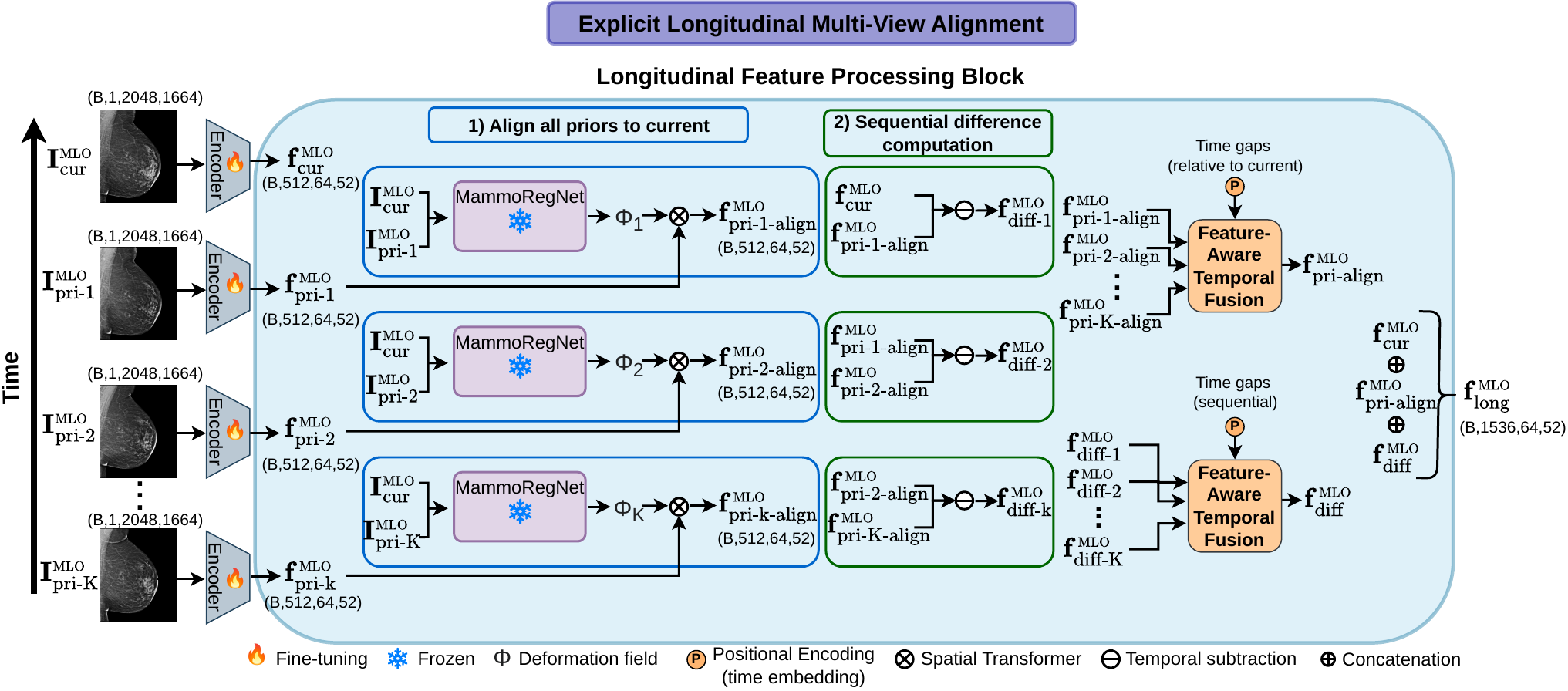}
    \caption{Overview of the explicit longitudinal multi-view feature alignment in FlexLMV-Net. For illustration, the alignment process is shown for the MLO view, where prior feature maps are aligned to the current features using image-derived deformation fields, enabling consistent correspondence and the construction of view-specific longitudinal representations. The CC view is processed in exactly the same manner. }
    \label{fig:lmvnet1}
\end{figure}

\noindent The explicit longitudinal alignment component of FlexLMV-Net, illustrated in Figure~\ref{fig:lmvnet1}, consists of a shared image encoder that processes longitudinal multi-view mammographic inputs, together with a longitudinal feature processing block designed to accommodate a variable number of prior examinations.

First, a shared encoder extracts feature maps from the current and available prior mammograms in both MLO and CC views. For a given patient, the current examination is represented by feature maps \(\mathbf{f}_{\text{cur}}^{\scriptscriptstyle \mathrm{MLO}}, \mathbf{f}_{\text{cur}}^{\scriptscriptstyle \mathrm{CC}} \in \mathbb{R}^{f_C \times f_H \times f_W}\), while the $(K)$ available prior examinations are represented by feature maps \(\mathbf{f}_{\text{pri-k}}^{\scriptscriptstyle \mathrm{MLO}}, \mathbf{f}_{\text{pri-k}}^{\scriptscriptstyle \mathrm{CC}}  \in \mathbb{R}^{f_C \times f_H \times f_W}\), for $k\in {(1, \ldots, K)}$, where $(K)$ may vary between patients.

Our approach leverages explicit longitudinal feature alignment to establish spatial correspondence between the current examination and each available prior examination, enabling anatomically consistent modeling of temporal changes. Specifically, we use image-derived deformation fields obtained from a pretrained and frozen MammoRegNet~\citep{mammoregnet} to align each prior examination to the current examination. For each prior examination and view, MammoRegNet produces a deformation field  \(\boldsymbol{\phi} \in \mathbb{R}^{2 \times H \times W}\), which is downsampled to \(\boldsymbol{\phi} \in \mathbb{R}^{2 \times f_H \times f_W}\) to match the feature-map resolution. The corresponding deformation field is then applied to the prior feature map using a Spatial Transformer Block~\citep{nicetrans,mammoregnet} with bilinear sampling, yielding aligned prior features \(\mathbf{f}_{\text{pri-k-align}}^{\scriptscriptstyle \mathrm{MLO}}, \mathbf{f}_{\text{pri-k-align}}^{\scriptscriptstyle \mathrm{CC}}\) for $k\in {(1, \ldots, K)}$. This process is performed independently for each available prior examination, resulting in a sequence of spatially aligned longitudinal representations for each view.

To capture temporal changes, view-specific difference features are computed between consecutive examinations using the aligned prior features. For the most recent prior examination, the difference is computed with respect to the current examination, while differences between earlier examinations are computed sequentially. Specifically, for each view, the resulting difference features capture changes between temporally adjacent examinations after spatial alignment. 

The resulting sequences of aligned prior and difference features can contain a variable number of elements depending on the available longitudinal imaging history. Rather than requiring a fixed number of prior examinations, FlexLMV-Net uses a Feature-Aware Temporal Fusion block to independently aggregate the aligned prior and difference features into compact longitudinal representations. The temporal gaps between screenings are incorporated into this fusion process to account for varying intervals between examinations. The resulting fused representations are then combined with the current features to form the longitudinal CC and MLO representations, which are subsequently processed by the multi-view dual-stream attention block.

The Feature-Aware Temporal Fusion block and its handling of variable-length longitudinal sequences are described in detail in the following subsection.

\subsection{Feature-Aware Temporal Fusion}
\label{feature_aware_temporal_fusion}

\begin{figure}
    \centering
    \includegraphics[width=0.85\linewidth]{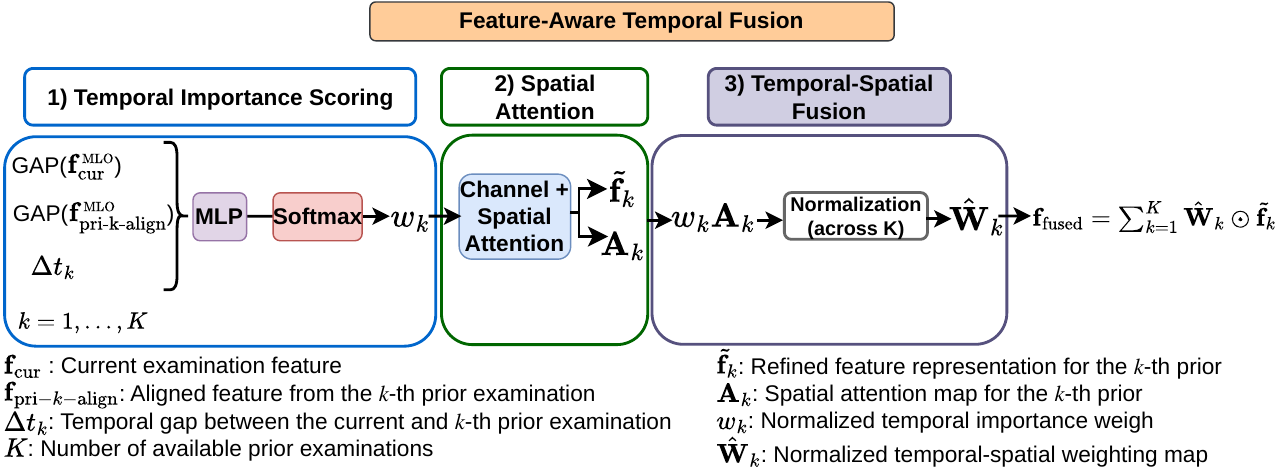}
    \caption{The Feature-aware temporal fusion of FlexLMV-Net adaptively weights longitudinal features based on temporal importance and spatial relevance.}
    \label{fig:feature_aware_fusion}
\end{figure}

To flexibly aggregate longitudinal information from a variable number of prior examinations, we introduce a Feature-Aware Temporal Fusion block (Figure~\ref{fig:feature_aware_fusion}). The block independently fuses the aligned prior features and difference features, allowing the model to adaptively determine the contribution of each available examination based on its feature content, temporal distance, and spatial relevance. Let $K$ denote the number of available prior examinations for a patient, with $k=1,\ldots,K$ indexing the individual prior examinations.

For each longitudinal feature $\mathbf{f}_k$, the pooled current feature, pooled longitudinal feature, and absolute temporal gap $\Delta t_k$ are concatenated and passed through a lightweight multilayer perceptron (MLP) to obtain a temporal importance score:

$$
s_k =
\mathrm{MLP}\left(
[\mathrm{GAP}(\mathbf{f}_{\mathrm{cur}});
\mathrm{GAP}(\mathbf{f}_k);
\Delta t_k]
\right),
\qquad k=1,\ldots,K.
$$
The MLP consists of two linear layers with a ReLU activation and dropout between them, producing a scalar importance score for each prior examination. The resulting scores are normalized across the $K$ available examinations using a softmax:

$$
w_k =
\frac{\exp(s_k)}
{\sum_{j=1}^{K}\exp(s_j)},
\qquad k=1,\ldots,K,
$$

where $s_k$ denotes the scalar importance score assigned to the $k$-th prior examination, and $w_k$ denotes its normalized temporal importance weight. The input to the MLP has dimension $2f_C+1$, corresponding to the pooled current and longitudinal features together with the temporal gap. Dropout provides regularization of the learned temporal weighting.

In parallel, a spatial attention module applies channel and spatial attention to each longitudinal feature, producing a refined feature representation and a corresponding spatial attention map. The temporal weight is then combined with the spatial attention map to obtain a joint temporal-spatial weighting map:

$$
\mathbf{W}_k = w_k \mathbf{A}_k,
\qquad k=1,\ldots,K.
$$

where $\mathbf{A}_k$ denotes the spatial attention map for the $k$-th prior examination and $\mathbf{W}_k$ denotes the resulting unnormalized temporal-spatial weighting map. These weighting maps are subsequently normalized across the $K$ available examinations independently at each spatial location, yielding $\hat{\mathbf{W}}_k$. The refined longitudinal features are then fused using the resulting spatially varying weights:

$$
\mathbf{f}_{\mathrm{fused}}
=
\sum_{k=1}^{K}
\hat{\mathbf{W}}_k \odot \tilde{\mathbf{f}}_k,
$$

where $\tilde{\mathbf{f}}_k$ denotes the refined longitudinal feature representation of the $k$-th prior examination, $\hat{\mathbf{W}}_k$ denotes its normalized temporal-spatial weighting map, and $\odot$ denotes element-wise multiplication.

This feature-aware fusion enables FlexLMV-Net to selectively emphasize both informative examinations and relevant spatial regions while naturally accommodating a variable number of available priors. The fusion process is applied independently to the aligned prior features and the difference features, which are subsequently combined with the current features to form the longitudinal CC and MLO representations.

\subsection{Multi-View Dual-Stream Attention}
\begin{figure}[tbp]
    \centering
    \includegraphics[width=0.95\linewidth]{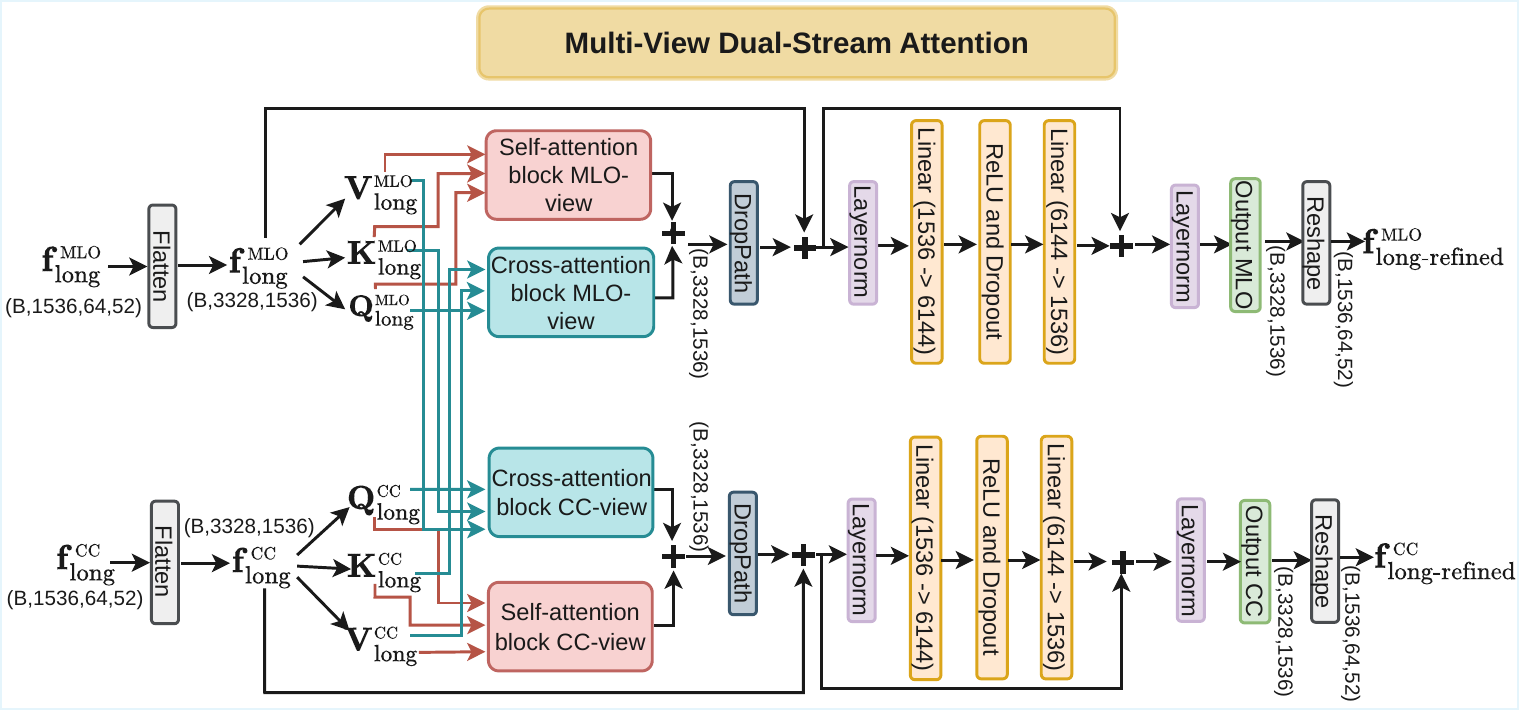}
    \caption{Overview of the Multi-View Dual-Stream Attention block in FlexLMV-Net. Longitudinal features from the MLO and CC views are tokenized and processed through parallel self- and cross-attention streams to model intra- and inter-view dependencies. The resulting features are then refined via residual connections and reshaped back into spatial feature maps, yielding enhanced multi-view representations.}
    \label{fig:lmvnet2}
\end{figure}

\noindent We propose a Multi-view Dual-Stream Attention block (see Figure~\ref{fig:lmvnet2}) to enhance the modeling of longitudinal multi-view representations by jointly refining the feature maps $\mathbf{f}_{\text{long}}^{\scriptscriptstyle \mathrm{MLO}}\in \mathbb{R}^{3f_C \times f_H \times f_W}$ and $ \mathbf{f}_{\text{long}}^{\scriptscriptstyle \mathrm{CC}} \in \mathbb{R}^{3f_C \times f_H \times f_W}$. The block operates on flattened spatial features and processes each view through two parallel attention streams: a self-attention stream and a cross-attention stream.

For each view, the input feature maps $\mathbf{f}_{\text{long}}\in \mathbb{R}^{3f_C \times f_H \times f_W}$ are first reshaped into a sequence of tokens $\in \mathbb{R}^{(f_H \cdot f_W) \times 3f_C}$. Queries, keys, and values are then obtained via linear projections. In the self-attention stream, queries, keys, and values originate from the same view (e.g., MLO), enabling the model to capture long-range intra-view dependencies and improve spatial coherence. In parallel, the cross-attention stream uses queries from one view and keys and values from the complementary view (e.g., CC), allowing the model to attend to anatomically and semantically corresponding regions across views.

The outputs of the self-attention and cross-attention streams are fused via element-wise summation, producing a unified attention representation for each view. DropPath~\citep{droppath} regularization is applied to the fused features to improve generalization. The result is then combined with the original input through a residual connection, followed by layer normalization~\citep{layernorm}.
Each view is subsequently processed by a feed-forward network (FFN), consisting of two linear layers with ReLU activation and dropout~\citep{dropout}. The FFN expands the feature dimension by a factor of four before projecting it back to the original size, increasing representational capacity. A second residual connection and layer normalization are applied after the FFN.
Finally, the refined token sequences are reshaped back to their original spatial dimensions, yielding updated feature maps $\mathbf{f}_{\text{long-refined}}^{\scriptscriptstyle \mathrm{MLO}}, \mathbf{f}_{\text{long-refined}}^{\scriptscriptstyle \mathrm{CC}} \in \mathbb{R}^{f_C \times f_H \times f_W}$ for each view. Through this dual-stream attention design, the model effectively integrates intra-view spatial context and inter-view complementary information, resulting in enriched and discriminative multi-view representations for downstream risk prediction.

\subsection{Risk Prediction Head}

\begin{figure}[tb]
    \centering
    \includegraphics[width=0.6\linewidth]{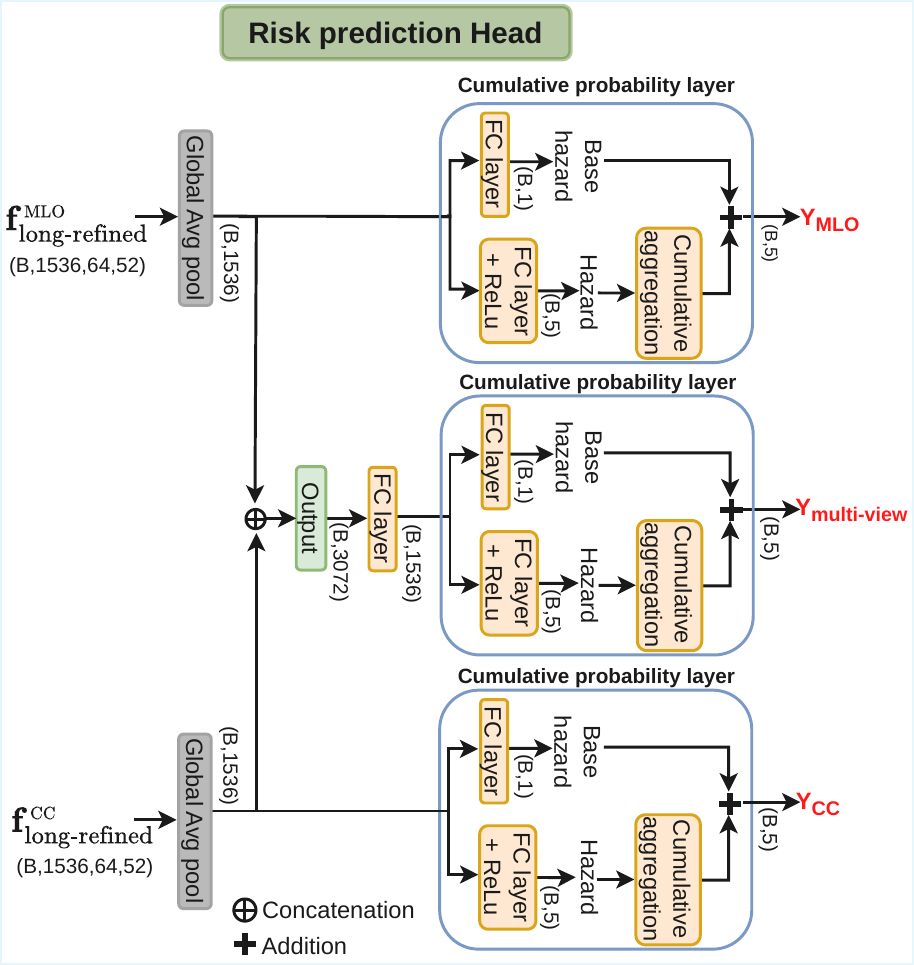}
    \caption{Risk prediction head of FlexLMV-Net. Refined MLO and CC features are globally pooled and fused for risk estimation via a cumulative probability layer, with auxiliary view-specific branches providing additional supervision during training.
 }
    \label{fig:lmvnet3}
\end{figure}

\noindent Figure~\ref{fig:lmvnet3} illustrates the risk prediction head, which takes the refined longitudinal feature maps \(\mathbf{f}_{\text{long-refined}}^{\scriptscriptstyle \mathrm{MLO}}\) and \(\mathbf{f}_{\text{long-refined}}^{\scriptscriptstyle \mathrm{CC}}\) as input to estimate time-dependent breast cancer risk.
The refined feature maps  \(\mathbf{f}_{\text{long-refined}}^{\scriptscriptstyle \mathrm{MLO}}\) and \(\mathbf{f}_{\text{long-refined}}^{\scriptscriptstyle \mathrm{CC}}\) are first globally pooled to obtain compact feature vectors for each view. These are then concatenated to form a fused multi-view representation. In addition to the fused representation, view-specific branches are retained to provide auxiliary supervision during training.
Risk estimation is performed using a cumulative probability layer~\citep{mirai,primeplus}, a formulation commonly used in survival analysis for time-to-event modeling. This layer consists of two fully connected branches: one predicting hazard values $h_j(\mathbf{f})$ at each future time step $j \in {1, \dots, T_{\text{max}}}$, and another estimating a baseline hazard $h_{\text{base}}(\mathbf{f})$. A ReLU activation function constrains the predicted hazard values to be non-negative. The cumulative risk score up to time $t$ is then computed as
\begin{equation}
z(t \mid \mathbf{f}) = h_{\text{base}}(\mathbf{f}) + \sum_{j=1}^{t} h_j(\mathbf{f}),
\end{equation}
which models risk as a sequential accumulation of hazards over time. During training, the cumulative risk scores are passed to the binary cross-entropy loss with logits, which applies a sigmoid transformation to map the scores to probabilities in the range $[0,1]$.
To encourage both global and view-specific learning, we adopt a multilevel joint learning strategy~\citep{Wan_Ordinal_MICCAI2024}. The fused multi-view representation is used for the primary risk prediction, while the individual MLO and CC feature vectors are passed through separate cumulative probability layers to produce auxiliary predictions. This setup promotes discriminative feature learning at the view level while improving the quality of the fused representation used at inference.
The model is trained using a time-dependent binary cross-entropy (BCE) loss with logits. The ground-truth $\mathbf{y}$ is a binary vector of length $T_{\text{max}}$ and encodes whether cancer occurs within each time horizon $t$, with $T_{\text{max}} = 5$ years in our setting. Specifically, $y(t) = 1$ if cancer is diagnosed within $t$ years, and $0$ otherwise. The sigmoid transformation inherent to the BCE-with-logits formulation converts the cumulative risk scores into probabilities for optimization. To handle censored data, we employ a masking function $\delta(t)$~\citep{mirai,Wan_Ordinal_MICCAI2024}, which ensures that only valid time points contribute to the loss:
\begin{equation}\label{mask}
    \delta(t) =
\begin{cases} 
        1, & \begin{aligned}[t] 
                & \text{if the follow-up period} \geq \min(t, T_{max}) \\  
                & \text{or if the patient is diagnosed within } t \text{ years}.
            \end{aligned} \\ 
        0, & \text{otherwise}.
    \end{cases}
\end{equation}

\noindent The overall loss is computed as
\begin{equation}
    \mathcal{L}_{\text{BCE}} = \sum_{t=1}^{T_{\text{max}}} \delta(t) \cdot \ell_{\text{BCE}}(t) \; ,
\end{equation}
where the per-time-step binary cross-entropy is defined as
\begin{equation}
    \ell_{\text{BCE}}(t) = -\left[  y(t) \cdot \log(P(t)) + (1 -  y(t) \cdot \log(1 -P(t))) \right] \; ,
\end{equation}
This formulation ensures that loss contributions are restricted to observed and clinically valid time points, enabling robust learning in the presence of censored longitudinal data.

\section{Experimental Setup and BCRBench}\label{expsetup}
This section outlines the experimental setup for training and evaluating our models. We describe the datasets, label generation and preprocessing (Section~\ref{datasetshort}), evaluation metrics (Section~\ref{evalmet}), and model implementation and training details (Section~\ref{impdet}). Our proposed BCRBench is described in Section~\ref{bcrbenc}.

\subsection{Datasets, Label Generation, and Preprocessing}\label{datasetshort}

\subsubsection{Datasets}
We evaluate our method on two large-scale public available longitudinal mammography datasets: the Emory Breast Imaging Dataset (EMBED)\footnote{\url{https://aws.amazon.com/marketplace/pp/prodview-unw4li5rkivs2\#overview}}~\citep{embed}
 and the Cohort of Screen-Aged Women Case-Control (CSAW-CC) dataset\footnote{\url{https://snd.se/en/catalogue/dataset/2021-204-1}}~\citep{csawcc}. Both datasets consist of two-dimensional (2D) full-field digital mammography (FFDM) images collected over multiple years, enabling comprehensive longitudinal analysis. Synthetic 2D mammograms (C-view images) from digital breast tomosynthesis (DBT) were excluded.

The publicly available EMBED dataset contains 364,564 mammograms from 23,256 patients acquired between 2013 and 2020 using the vendors Hologic, GE and Fujifilm. CSAW-CC includes mammograms from 873 breast cancer cases and 7,850 controls screened between 2008 and 2015, all acquired using Hologic equipment. Both datasets include invasive and non-invasive breast cancer cases. Negative cases are defined as women without a breast cancer diagnosis during the respective follow-up periods (8 years for EMBED and 7 years for CSAW-CC).

To ensure sufficient longitudinal information, only patients with at least five years of follow-up were included, following prior work~\citep{Wan_Ordinal_MICCAI2024}. For patients with more than two screening exams, all consecutive screening pairs were used for training.

\subsubsection{Time-to-Events label generation}
The Time-to-Events label represents the temporal interval between a screening mammogram and the corresponding breast cancer diagnosis. Labels were generated following established protocols~\citep{Wan_Ordinal_MICCAI2024,Kar_Longitudinal_MICCAI2024,thrunarxiv2025}, with dataset-specific procedures for identifying positive and negative cases and assigning cancer timing.

\textbf{EMBED:} Labeling follows \cite{Wan_Ordinal_MICCAI2024,thrunarxiv2025}. Positive cases include mammograms with BI-RADS 6 (biopsy-proven malignancy) or pathological severity scores indicating invasive or non-invasive cancer. Negative cases include BI-RADS 1–2 exams, as well as BI-RADS 0 exams that were subsequently resolved as BI-RADS 1 or 2 in follow-up studies. The cancer diagnosis date is defined as the biopsy confirmation date.

\textbf{CSAW-CC:} Labeling follows \cite{Kar_Longitudinal_MICCAI2024,thrunarxiv2025}, using the radiological timing variable (rad\_timing). Positive cases correspond to screen-detected cancers (rad\_timing = 1) and interval cancers (rad\_timing = 2), while negative cases consist of exams without any recorded rad\_timing, indicating no cancer diagnosis during the study period. Since biopsy dates are not available in CSAW-CC, cancer timing is inferred from rad\_timing. For each patient, the cancer year is determined from the most recent exam: it is set to the exam year for screen-detected cancers and to one year after the exam for interval cancers. For cases in 2016 (the final study year), the cancer year is set to 2016 to accommodate follow-up constraints.

\textbf{Time-to-Events:} For both datasets, the Time-to-Events label is calculated as the difference between the assigned cancer year and the mammogram exam year. Only negative prior screening mammograms are included, as routine screening usually stops once a patient is diagnosed with breast cancer.

\subsubsection{Preprocessing}
All mammograms were preprocessed following established pipelines~\citep{mirai}. Raw DICOM images were converted to 16-bit grayscale PNG format, after which background artifacts were removed and the breast region was segmented to exclude non-breast areas. The resulting images were resized to a standardized resolution of \(1664 \times 2048\) pixels and intensity-normalized to ensure consistency across scanners and acquisition settings. Density information is derived from BI-RADS~\citep{birads} categories for EMBED, while for CSAW-CC, LIBRA-based~\citep{libra} percent density estimates are grouped into three categories of equal size corresponding to low, medium and high density. Detailed statistics on the Time-to-Cancer distribution and dataset splits are provided in the Appendix in Table~\ref{tab:timecancerdataset}, while breast density distributions are summarized in Table~\ref{tab:combined_densitydistr}.

\subsection{Evaluation Metrics}\label{evalmet}

Following previous studies~\citep{primeplus,Wan_Ordinal_MICCAI2024,mirai}, we evaluated the performance of the methods on the 1–5 year risk prediction task using the concordance index (C-index)~\citep{c_index} and the area under the receiver operating characteristic curve (AUC)~\citep{auc} across the 5 year risk horizons. 95\% confidence intervals (CIs) were estimated via bootstrapping with 1,000 resamples. Statistical significance of improvements over the second-best model was assessed using a paired nonparametric bootstrap test~\citep{kang2015cindex,Wang2023}, with $p<0.05$ considered significant.

\subsection{Implementation Details}\label{impdet}
All models were implemented in PyTorch 2.0.1~\citep{pytorch} and trained for up to 30 epochs on eight AMD Instinct MI210 GPUs (64 GB each), using a batch size of 4 per GPU (32 total). We compare FlexLMV-Net with Mirai~\citep{mirai}, VMRA-MaR~\citep{vmramar}, OA-BreaCR~\citep{Wan_Ordinal_MICCAI2024}, ImgFeatAlign~\citep{mammoregnet,thrunarxiv2025}, and LMV-Net~\citep{ownpaperMICCAI2026}. For fair comparison, we use the Mirai encoder~\citep{mirai}, a ResNet-18~\citep{resnet18} without the final fully connected and global average pooling layers, following prior work~\citep{Kar_Longitudinal_MICCAI2024,vmramar,thrunarxiv2025}. Self- and cross-attention use four heads. Models are optimized with AdamW~\citep{adamw} using a learning rate of $5 \times 10^{-5}$ ($5 \times 10^{-6}$ for encoder fine-tuning) and weight decay of $1 \times 10^{-4}$. The learning rate is halved after five epochs without validation C-index improvement, with early stopping after 15 epochs. Augmentation includes random cropping, affine transformations, color jitter, and gamma correction. Comparison methods use their official configurations. LMV-Net is evaluated with both frozen and fine-tuned Mirai encoders, whereas FlexLMV-Net uses the fine-tuned encoder, which yielded improved performance for LMV-Net. In this study, FlexLMV-Net considers up to three prior examinations due to memory limitations.

\subsection{BCRBench}\label{bcrbenc}
As part of this study, we developed BCRBench (\url{https://github.com/sot176/BCRBench}), an open-source benchmarking framework for image-based breast cancer risk prediction. BCRBench addresses the lack of standardized experimental setups by unifying preprocessing, data handling, training, and evaluation within a reproducible framework, enabling consistent comparisons across models and datasets.

Implemented in PyTorch, BCRBench follows a modular design through which datasets, models, and evaluation protocols can be integrated via a common interface. The current version includes EMBED and CSAW-CC and five risk prediction models: Mirai~\citep{mirai}, VMRA-MaR~\citep{vmramar}, OA-BreaCR~\citep{Wan_Ordinal_MICCAI2024}, ImgFeatAlign~\citep{mammoregnet,thrunarxiv2025}, LMV-Net~\citep{ownpaperMICCAI2026}, and FlexLMV-Net. These models use standard CC and MLO views and cover single-timepoint and longitudinal risk prediction. For Mirai-based models, including VMRA-MaR, ImgFeatAlign, LMV-Net, and FlexLMV-Net, both frozen and end-to-end fine-tuned encoder settings are supported.

BCRBench applies consistent preprocessing, training, and evaluation procedures across methods. Performance is assessed using AUC and C-index across five-year prediction horizons, with support for subgroup analyses based on clinically relevant characteristics such as breast density and cancer subtype. The framework provides the experimental foundation for the comparisons presented in this study and is openly available with comprehensive documentation (\url{https://sot176.github.io/BCRBench}) to facilitate reproducibility and future benchmarking.

\section{Results}\label{result}
This section presents the quantitative and qualitative results of FlexLMV-Net and its special case LMV-Net, where $K=1$, for five-year breast cancer risk prediction. Section~\ref{sotaresults} compares both models with SOTA methods, while Sections~\ref{performdensity}--\ref{subgroups} evaluate performance across breast density, cancer type, and demographic subgroups. Section~\ref{generalperform} assesses in-domain and cross-dataset generalization, and Section~\ref{attentanalysis} analyzes the interpretability of the dual-stream attention block. Section~\ref{history} examines the effect of varying longitudinal history length, while Section~\ref{ablstudy} presents the ablation study.

\begin{table}[t]
\centering
\scriptsize
\caption{1–5 Year breast cancer risk prediction: C-index and AUC values with $\pm$ 95\% CI. \textcolor{blue}{\scalebox{0.9}{\faSnowflake}}: frozen encoder,  
\textcolor{orange}{\scalebox{0.9}{\faFire}}: fine-tuned encoder. $\Delta (\%)$ denotes difference between FlexLMV-Net and best performing baseline. $^{*}$ indicates statistically significant improvement of LMV-Net and FlexLMV-Net over the best baseline (p < 0.05). Training time in hours (h) and total number of model parameters (\#, in millions).}

\begin{adjustbox}{max width=\textwidth}
\begin{tabular}{clc*{5}{c}cc}
\toprule
\textbf{Dataset} & \textbf{Method} & \textbf{C-index (\%) $\uparrow$} & \multicolumn{5}{c}{\textbf{Follow-up year AUC (\%) $\uparrow$}} & \multirow{2}{*}{h} & \multirow{2}{*}{ \#} \\
\cmidrule(lr){4-8}
& & & \textbf{1-yr} & \textbf{2-yr} & \textbf{3-yr} & \textbf{4-yr} & \textbf{5-yr} \\
\midrule

\multirow{7}{*}{\rotatebox{90}{\textbf{EMBED}}}
  
    & Mirai(\textcolor{blue}{\scalebox{0.9}{\faSnowflake}})
& \makecell{59.8\fontsize{6}{6}\selectfont$\pm$8.7} 
& \makecell{56.7\fontsize{6}{6}\selectfont$\pm$20.7} 
& \makecell{52.6\fontsize{6}{6}\selectfont$\pm$15.3} 
& \makecell{54.0\fontsize{6}{6}\selectfont$\pm$12.6} 
& \makecell{49.3\fontsize{6}{6}\selectfont$\pm$12.5}
& \makecell{49.2\fontsize{6}{6}\selectfont$\pm$11.6} & 6& 28\\

    & Mirai(\textcolor{orange}{\scalebox{0.9}{\faFire}})
& \makecell{69.1\fontsize{6}{6}\selectfont$\pm$10.6} 
& \makecell{71.0\fontsize{6}{6}\selectfont$\pm$15.6} 
& \makecell{70.9\fontsize{6}{6}\selectfont$\pm$15.5}
& \makecell{70.6\fontsize{6}{6}\selectfont$\pm$11.7} 
& \makecell{65.7\fontsize{6}{6}\selectfont$\pm$11.3} 
& \makecell{65.9\fontsize{6}{6}\selectfont$\pm$11.6} & 7 & 28\\

    & VMRA-MaR (\textcolor{orange}{\scalebox{0.9}{\faFire}}) 
& \makecell{72.4\fontsize{6}{6}\selectfont$\pm$12.5} 
& \makecell{77.9\fontsize{6}{6}\selectfont$\pm$4.5} 
& \makecell{75.3\fontsize{6}{6}\selectfont$\pm$9.7}
& \makecell{73.1\fontsize{6}{6}\selectfont$\pm$12.7}
& \makecell{68.9\fontsize{6}{6}\selectfont$\pm$14.3} 
& \makecell{67.1\fontsize{6}{6}\selectfont$\pm$11.8} & 45 & 2802\\

  & OA-BreaCR(\textcolor{orange}{\scalebox{0.9}{\faFire}}) 
  & \makecell{72.3\fontsize{6}{6}\selectfont$\pm$2.0}
  & \makecell{72.9\fontsize{6}{6}\selectfont$\pm$2.8}
  & \makecell{72.3\fontsize{6}{6}\selectfont$\pm$4.9}
  & \makecell{72.4\fontsize{6}{6}\selectfont$\pm$2.5}
  & \makecell{72.3\fontsize{6}{6}\selectfont$\pm$2.4} 
                      & \makecell{73.2\fontsize{6}{6}\selectfont$\pm$2.4} & 9& 15\\

  & ImgFeatAlign (\textcolor{blue}{\scalebox{0.9}{\faSnowflake}}) 
  & \makecell{74.7\fontsize{6}{6}\selectfont$\pm$2.4}
  & \makecell{75.0\fontsize{6}{6}\selectfont$\pm$2.9} 
  & \makecell{75.5\fontsize{6}{6}\selectfont$\pm$2.4} 
  & \makecell{75.3\fontsize{6}{6}\selectfont$\pm$2.1} 
  & \makecell{75.9\fontsize{6}{6}\selectfont$\pm$2.3} 
  & \makecell{72.5\fontsize{6}{6}\selectfont$\pm$3.2} &12 & 52 \\
  
& ImgFeatAlign (\textcolor{orange}{\scalebox{0.9}{\faFire}})
&   \makecell{72.2\fontsize{6}{6}\selectfont$\pm$2.7}    
&   \makecell{72.1\fontsize{6}{6}\selectfont$\pm$3.0}   
&   \makecell{73.3\fontsize{6}{6}\selectfont$\pm$2.6}   
&   \makecell{72.2\fontsize{6}{6}\selectfont$\pm$2.5}   
&   \makecell{71.3\fontsize{6}{6}\selectfont$\pm$2.4}  
&   \makecell{72.6\fontsize{6}{6}\selectfont$\pm$3.3} & 14 & 52 \\

\cdashline{2-10} 
& LMV-Net (\textcolor{blue}{\scalebox{0.9}{\faSnowflake}}) 
  & \makecell{78.3$^{*}$\fontsize{6}{6}\selectfont$\pm$3.0}
  & \makecell{80.8$^{*}$\fontsize{6}{6}\selectfont$\pm$3.7}
  & \makecell{78.3$^{*}$\fontsize{6}{6}\selectfont$\pm$3.5} 
  & \makecell{77.5$^{*}$\fontsize{6}{6}\selectfont$\pm$3.3} 
  & \makecell{76.6\fontsize{6}{6}\selectfont$\pm$3.2} 
  & \makecell{75.1$^{*}$\fontsize{6}{6}\selectfont$\pm$4.9} & 15 & 65\\

 &  LMV-Net (\textcolor{orange}{\scalebox{0.9}{\faFire}})  
& \makecell{\textbf{81.4$^{*}$}\textbf{\fontsize{6}{6}\selectfont $\pm$2.8}}
& \makecell{\textbf{83.9$^{*}$}\textbf{\fontsize{6}{6}\selectfont $\pm$3.4}}
& \makecell{\textbf{82.1$^{*}$}\textbf{\fontsize{6}{6}\selectfont $\pm$3.0}} 
& \makecell{\textbf{80.8$^{*}$}\textbf{\fontsize{6}{6}\selectfont $\pm$2.9}} 
& \makecell{\textbf{80.0$^{*}$}\textbf{\fontsize{6}{6}\selectfont $\pm$2.8}} 
& \makecell{{77.2$^{*}$}{\fontsize{6}{6}\selectfont $\pm$4.7}}  & 16 & 65\\
& FlexLMV-Net (\textcolor{orange}{\scalebox{0.9}{\faFire}})  
& \makecell{\textbf{81.2$^{*}$}\textbf{\fontsize{6}{6}\selectfont $\pm$2.9}}
& \makecell{\textbf{84.1$^{*}$}\textbf{\fontsize{6}{6}\selectfont $\pm$3.2}}
& \makecell{\textbf{82.0$^{*}$}\textbf{\fontsize{6}{6}\selectfont $\pm$3.0}} 
& \makecell{\textbf{79.6$^{*}$}\textbf{\fontsize{6}{6}\selectfont $\pm$3.1}} 
& \makecell{\textbf{79.3$^{*}$}\textbf{\fontsize{6}{6}\selectfont $\pm$3.1}} 
& \makecell{\textbf{79.0$^{*}$}\textbf{\fontsize{6}{6}\selectfont $\pm$4.4}} & 22 & 65.14\\
\hline
&  $\Delta$ (\%) $\uparrow$
& \makecell{\textbf{\textcolor{green!60!black}{+ 6.5}} }
& \makecell{\textbf{\textcolor{green!60!black}{+ 6.2}} }
& \makecell{\textbf{\textcolor{green!60!black}{+ 6.5}}} 
& \makecell{\textbf{\textcolor{green!60!black}{+ 4.3}}} 
& \makecell{\textbf{\textcolor{green!60!black}{+ 3.4}}} 
& \makecell{\textbf{\textcolor{green!60!black}{+ 5.8}}} & & \\
\midrule

\multirow{7}{*}{\rotatebox{90}{\textbf{CSAW-CC}}}

&  Mirai(\textcolor{blue}{\scalebox{0.9}{\faSnowflake}}) 
& \makecell{71.4\fontsize{6}{6}\selectfont$\pm$2.7} 
& \makecell{56.8\fontsize{6}{6}\selectfont$\pm$10.6} 
& \makecell{54.7\fontsize{6}{6}\selectfont$\pm$5.8} 
& \makecell{56.4\fontsize{6}{6}\selectfont$\pm$5.1} 
& \makecell{56.1\fontsize{6}{6}\selectfont$\pm$4.2} 
& \makecell{55.4\fontsize{6}{6}\selectfont$\pm$4.0} & 7 & 28\\

    &  Mirai (\textcolor{orange}{\scalebox{0.9}{\faFire}}) 
& \makecell{72.3\fontsize{6}{6}\selectfont$\pm$2.9} 
& \makecell{69.9\fontsize{6}{6}\selectfont$\pm$3.3} 
& \makecell{69.8\fontsize{6}{6}\selectfont$\pm$3.9}
& \makecell{69.2\fontsize{6}{6}\selectfont$\pm$3.8} 
& \makecell{67.0\fontsize{6}{6}\selectfont$\pm$3.4} 
& \makecell{66.9\fontsize{6}{6}\selectfont$\pm$3.4} &8 & 28\\

    &  VMRA-MaR (\textcolor{orange}{\scalebox{0.9}{\faFire}})  
& \makecell{72.7\fontsize{6}{6}\selectfont$\pm$2.6} 
& \makecell{70.9\fontsize{6}{6}\selectfont$\pm$5.5}
& \makecell{70.3\fontsize{6}{6}\selectfont$\pm$5.6} 
& \makecell{69.5\fontsize{6}{6}\selectfont$\pm$4.4} 
& \makecell{70.9\fontsize{6}{6}\selectfont$\pm$4.3} 
& \makecell{64.5\fontsize{6}{6}\selectfont$\pm$6.0} & 46& 2802\\

  &  OA-BreaCR (\textcolor{orange}{\scalebox{0.9}{\faFire}})  
& \makecell{61.6\fontsize{6}{6}\selectfont$\pm$4.2} 
& \makecell{63.5\fontsize{6}{6}\selectfont$\pm$4.4} 
& \makecell{61.7\fontsize{6}{6}\selectfont$\pm$2.1} 
& \makecell{65.0\fontsize{6}{6}\selectfont$\pm$2.0} 
& \makecell{67.0\fontsize{6}{6}\selectfont$\pm$1.9} 
& \makecell{67.7\fontsize{6}{6}\selectfont$\pm$2.2} & 10& 15\\

 & ImgFeatAlign (\textcolor{blue}{\scalebox{0.9}{\faSnowflake}})  
& \makecell{70.4\fontsize{6}{6}\selectfont$\pm$1.9} 
& \makecell{72.0\fontsize{6}{6}\selectfont$\pm$2.4} 
& \makecell{71.5\fontsize{6}{6}\selectfont$\pm$1.9} 
& \makecell{72.6\fontsize{6}{6}\selectfont$\pm$1.9}
& \makecell{72.0\fontsize{6}{6}\selectfont$\pm$2.1} 
& \makecell{75.2\fontsize{6}{6}\selectfont$\pm$2.3} &13 & 52\\

&  ImgFeatAlign (\textcolor{orange}{\scalebox{0.9}{\faFire}})
&  \makecell{71.9\fontsize{6}{6}\selectfont$\pm$2.0}
&  \makecell{74.1\fontsize{6}{6}\selectfont$\pm$2.6}  
&  \makecell{72.3\fontsize{6}{6}\selectfont$\pm$2.0} 
&  \makecell{73.1\fontsize{6}{6}\selectfont$\pm$1.9}   
&  \makecell{72.1\fontsize{6}{6}\selectfont$\pm$2.0} 
&   \makecell{74.5\fontsize{6}{6}\selectfont$\pm$2.3} & 16 & 52 \\

\cdashline{2-10} 
&  LMV-Net (\textcolor{blue}{\scalebox{0.9}{\faSnowflake}}) 
& \makecell{73.4$^{*}$\fontsize{6}{6}\selectfont$\pm$2.5}  
& \makecell{77.5$^{*}$\fontsize{6}{6}\selectfont$\pm$3.3}  
& \makecell{73.5$^{*}$\fontsize{6}{6}\selectfont$\pm$2.6}
& \makecell{74.0\fontsize{6}{6}\selectfont$\pm$2.5}
& \makecell{73.1\fontsize{6}{6}\selectfont$\pm$2.6} 
& \makecell{\textbf{75.9}\textbf{\fontsize{6}{6}\selectfont$\pm$2.8}}
& 16 & 65\\

 & LMV-Net (\textcolor{orange}{\scalebox{0.9}{\faFire}})  
& \makecell{ {74.4$^{*}$}{\fontsize{6}{6}\selectfont$\pm$2.7}} 
& \makecell{ {78.7$^{*}$}{\fontsize{6}{6}\selectfont$\pm$3.6} }
& \makecell{\textbf{74.1$^{*}$}\textbf{\fontsize{6}{6}\selectfont$\pm$2.9} }
& \makecell{\textbf{74.3$^{*}$}\textbf{\fontsize{6}{6}\selectfont$\pm$2.7}} 
& \makecell{ \textbf{73.5}\textbf{\fontsize{6}{6}\selectfont$\pm$2.8}} 
& \makecell{75.0\fontsize{6}{6}\selectfont$\pm$3.1} & 17& 65\\

& FlexLMV-Net (\textcolor{orange}{\scalebox{0.9}{\faFire}})  
& \makecell{\textbf{77.1$^{*}$}\textbf{\fontsize{6}{6}\selectfont $\pm$2.5}}
& \makecell{\textbf{82.5$^{*}$}\textbf{\fontsize{6}{6}\selectfont $\pm$3.0}}
& \makecell{\textbf{74.6$^{*}$}\textbf{\fontsize{6}{6}\selectfont $\pm$3.0}} 
& \makecell{\textbf{73.7$^{*}$}\textbf{\fontsize{6}{6}\selectfont $\pm$2.9}} 
& \makecell{{71.9}{\fontsize{6}{6}\selectfont $\pm$3.1}} 
& \makecell{{72.3}{\fontsize{6}{6}\selectfont $\pm$3.3}} & 26 & 65.14\\
\hline

&  $\Delta$ (\%) $\uparrow$
& \makecell{\textbf{\textcolor{green!60!black}{+ 4.4}} }
& \makecell{\textbf{\textcolor{green!60!black}{+ 10.5}}} 
& \makecell{\textbf{\textcolor{green!60!black}{+ 3.1}} }
& \makecell{\textbf{\textcolor{green!60!black}{+ 1.1}}} 
& \makecell{\textbf{\textcolor{red!60!black}{-0.1}}} 
& \makecell{\textbf{\textcolor{red!60!black}{-2.9}}} & &  \\
\bottomrule

\end{tabular}

\end{adjustbox}
\label{tab:riskpredoverall}
\end{table}

\subsection{Comparison with State-of-the-Arts (SOTA)}\label{sotaresults}
We compare the proposed FlexLMV-Net and LMV-Net with state-of-the-art (SOTA) methods, including Mirai~\citep{mirai} (Sci. Transl. Med.’21), OA-BreaCR~\citep{Wan_Ordinal_MICCAI2024} (MICCAI’24), VMRA-MaR~\citep{vmramar} (MICCAI’25), and ImgFeatAlign~\citep{thrunarxiv2025} (MICCAI’25), on both datasets. Table~\ref{tab:riskpredoverall} summarizes the results for 1--5 year breast cancer risk prediction. Both LMV-Net and FlexLMV-Net consistently outperform the competing methods across the two datasets. On EMBED, which encompasses diverse scanner types and population demographics, LMV-Net achieves substantial improvements over SOTA methods, demonstrating robustness to heterogeneous data. Fine-tuning the Mirai encoder further improves LMV-Net performance, highlighting the benefit of adapting the pretrained representation to the risk prediction task. FlexLMV-Net further improves upon LMV-Net at several prediction horizons while maintaining strong performance across the full 1--5 year range. Similarly, on the CSAW-CC dataset, which is less diverse and primarily derived from a single scanner type and a predominantly Scandinavian population, LMV-Net achieves consistent performance gains. Overall, LMV-Net establishes strong predictive performance, while FlexLMV-Net extends the framework to accommodate a variable number of prior examinations and achieves the strongest overall C-index, with particularly strong short-term risk prediction.
Moreover, both LMV-Net(\textcolor{orange}{\scalebox{0.9}{\faFire}}) and FlexLMV-Net (\textcolor{orange}{\scalebox{0.9}{\faFire}}) exhibit relatively narrow confidence intervals, indicating reliable and stable performance across prediction horizons. The improvements over the best baseline are statistically significant ($p<0.05$). 

\subsection{Performance across density categories}\label{performdensity}

\begin{figure}[tbhp]
    \centering
    \includegraphics[width=0.8\linewidth]{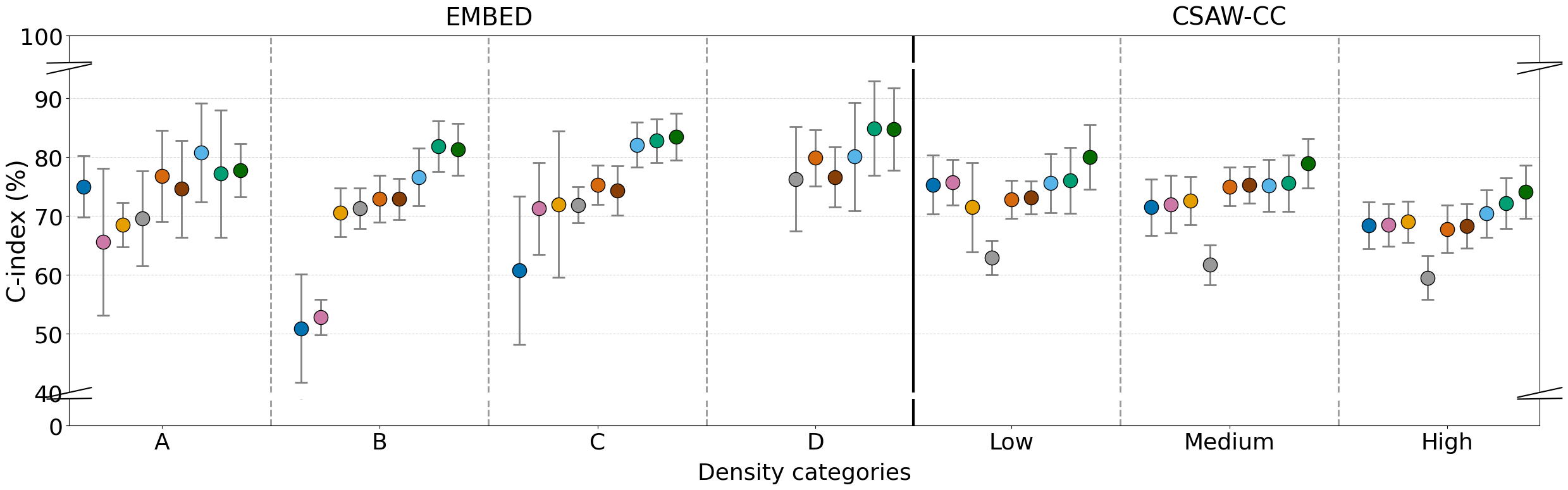} \\
\includegraphics[width=0.5\linewidth]{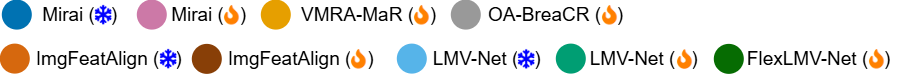}
    \caption{C-index across breast density categories for the proposed LMV-Net and state-of-the-art methods on the EMBED and CSAW-CC datasets (in-domain evaluation).
Points show the mean C-index and error bars represent the 95\% confidence interval.}
    \label{fig:cindexdensity}
\end{figure}

\begin{figure}[tbph]
\centering

\begin{subfigure}[t]{0.48\textwidth}
    \centering
    \includegraphics[width=\linewidth]{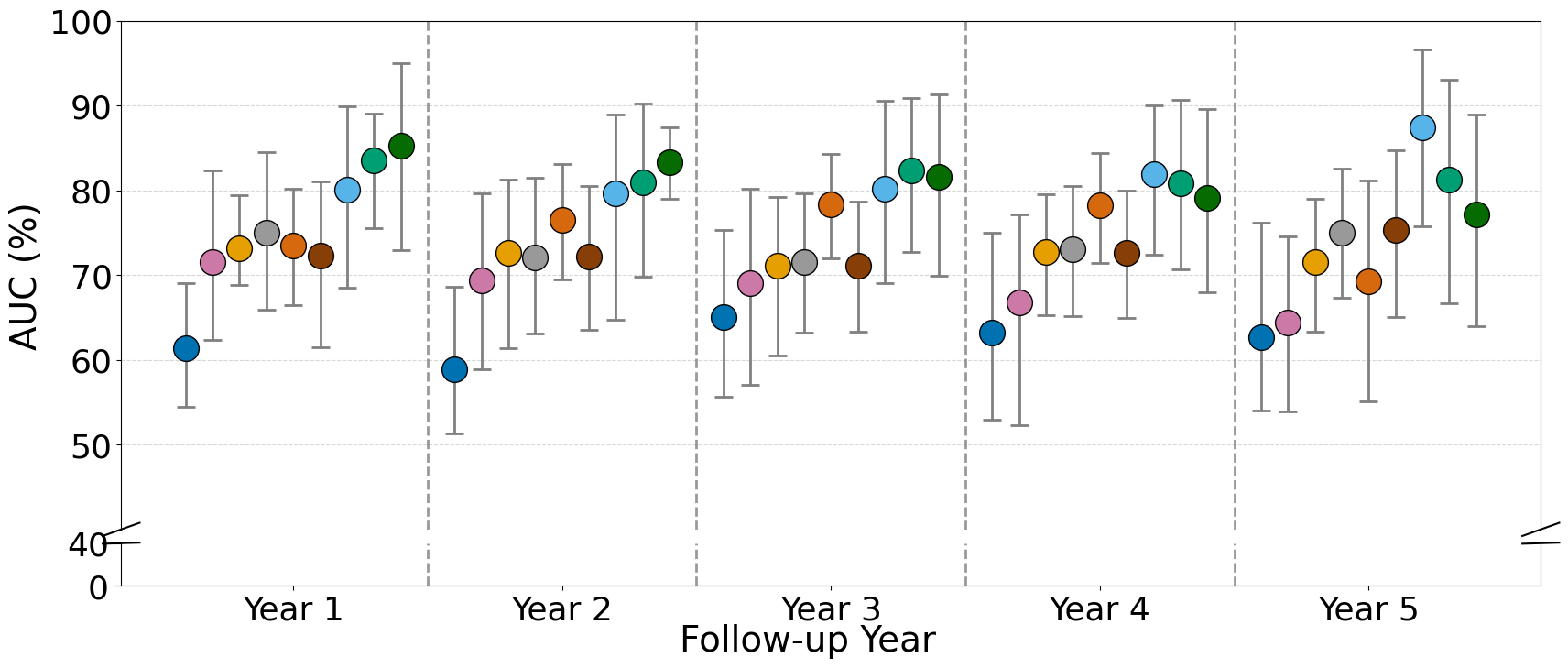}
    \caption{BI-RADS A}
\end{subfigure}
\quad
\begin{subfigure}[t]{0.48\textwidth}
    \centering
    \includegraphics[width=\linewidth]{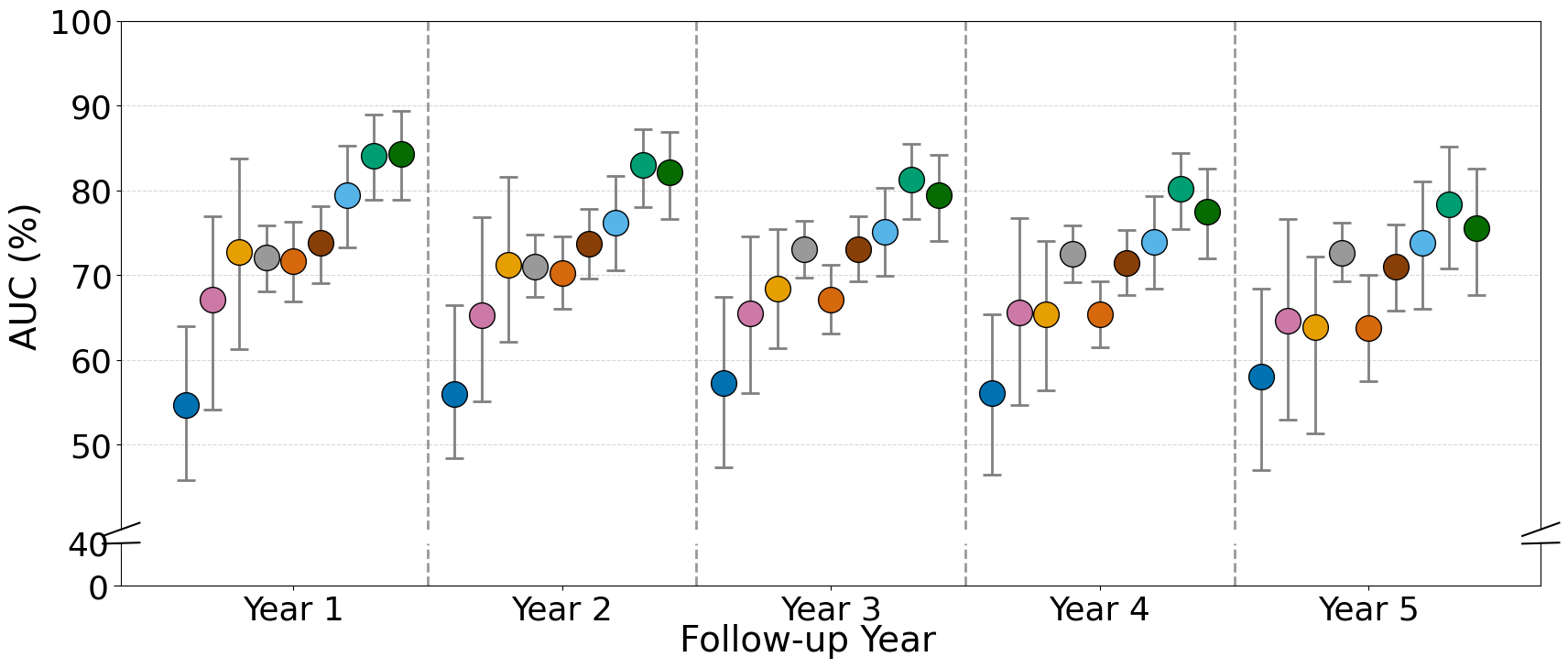}
    \caption{BI-RADS B}
\end{subfigure} \\

\begin{subfigure}[t]{0.48\textwidth}
    \centering
    \includegraphics[width=\linewidth]{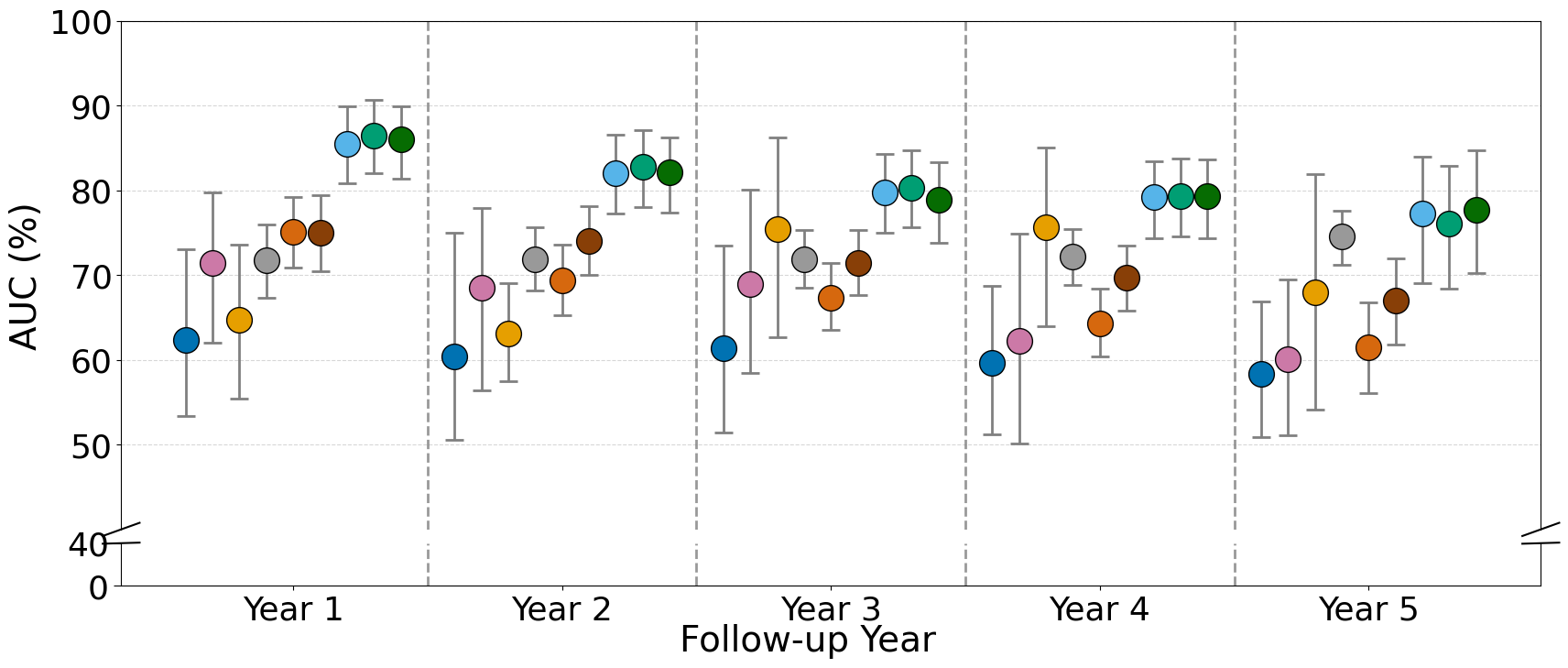}
    \caption{BI-RADS C}
\end{subfigure}
\quad
\begin{subfigure}[t]{0.48\textwidth}
    \centering
    \includegraphics[width=\linewidth]{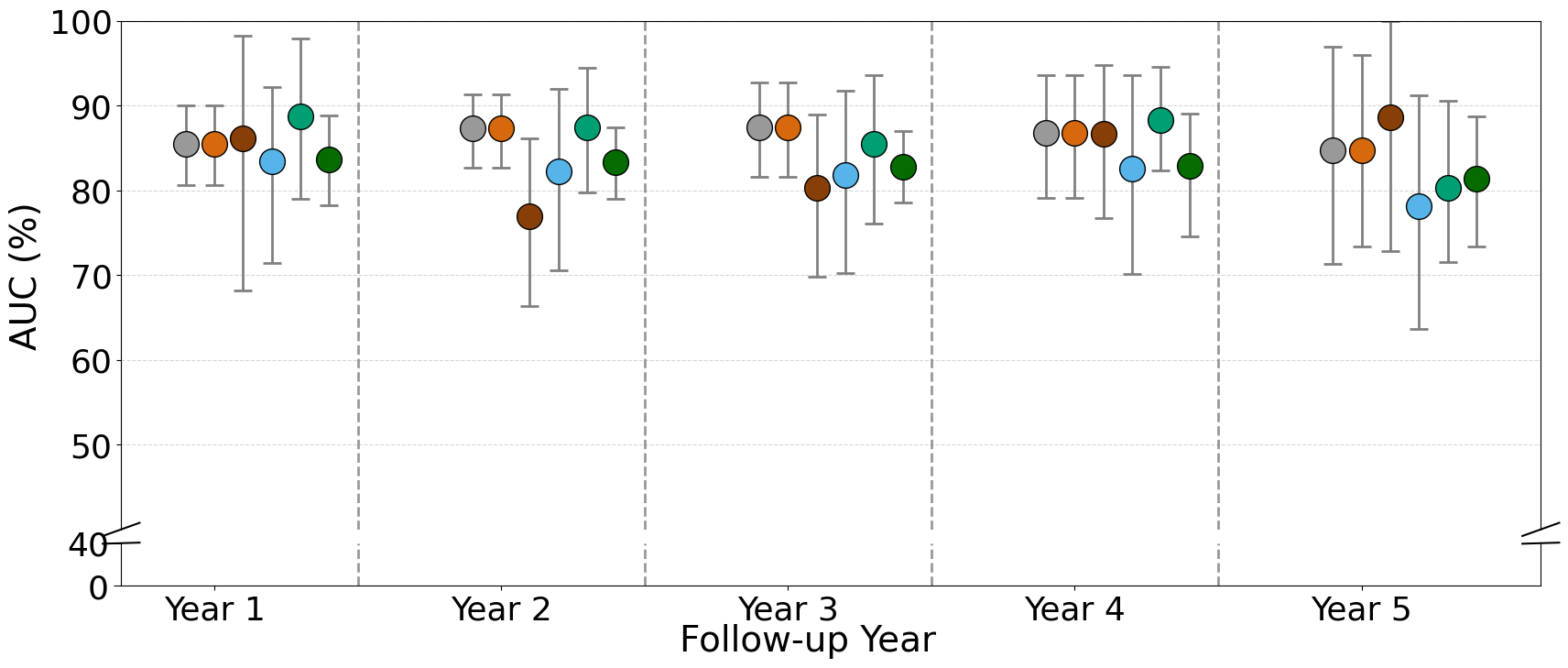}
    \caption{BI-RADS D}
\end{subfigure} 
 \\ \vspace{0.25cm}
\includegraphics[width=0.5\linewidth]{legend.png}

\caption{Forest plots showing AUC across five follow-up years for all models stratified by breast density categories (BI-RADS A-D) on the EMBED dataset. Points represent mean AUC and error bars represent 95\% confidence intervals.}
\label{fig:forest_density_auc_embed}
\end{figure}

\begin{figure}[tbph]
\centering

\begin{subfigure}[t]{0.48\textwidth}
    \centering
    \includegraphics[width=\linewidth]{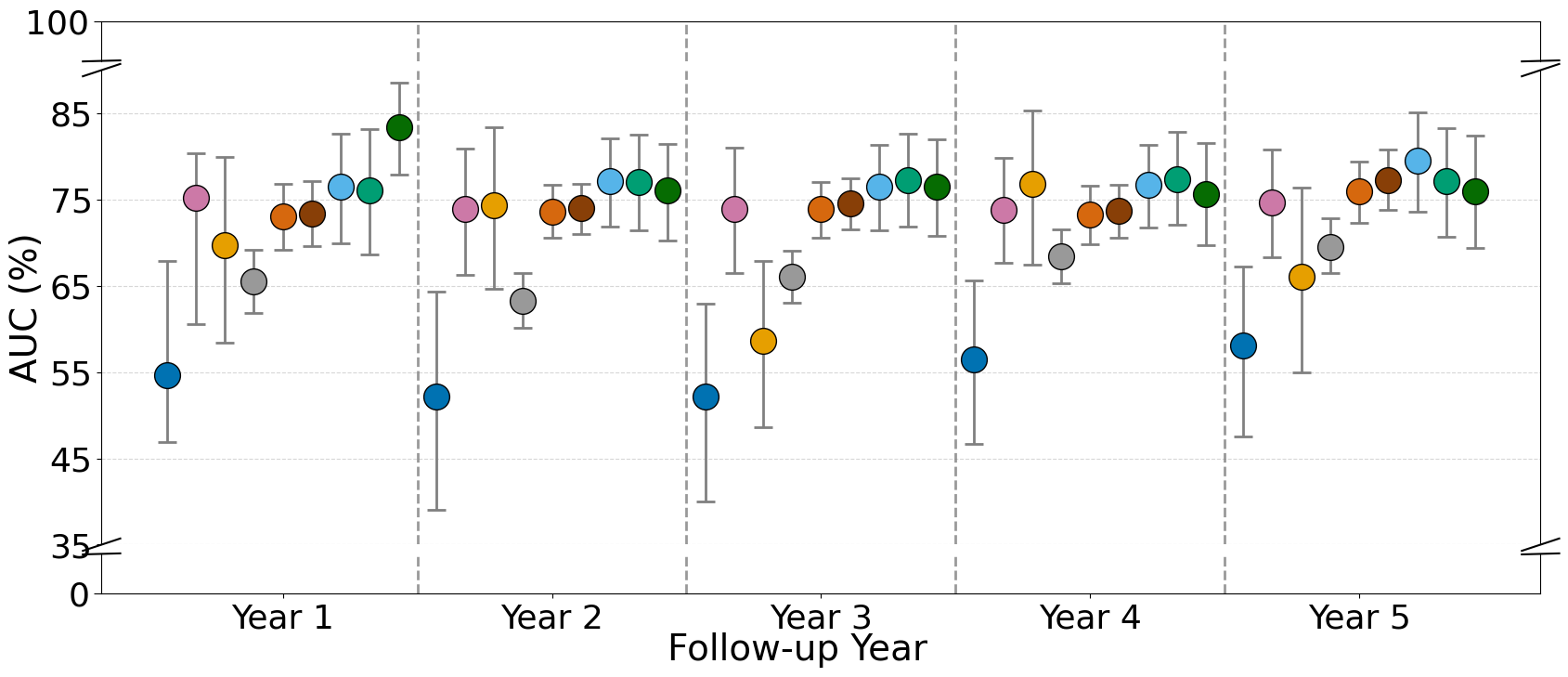}
    \caption{Low}
\end{subfigure}
\quad
\begin{subfigure}[t]{0.48\textwidth}
    \centering
    \includegraphics[width=\linewidth]{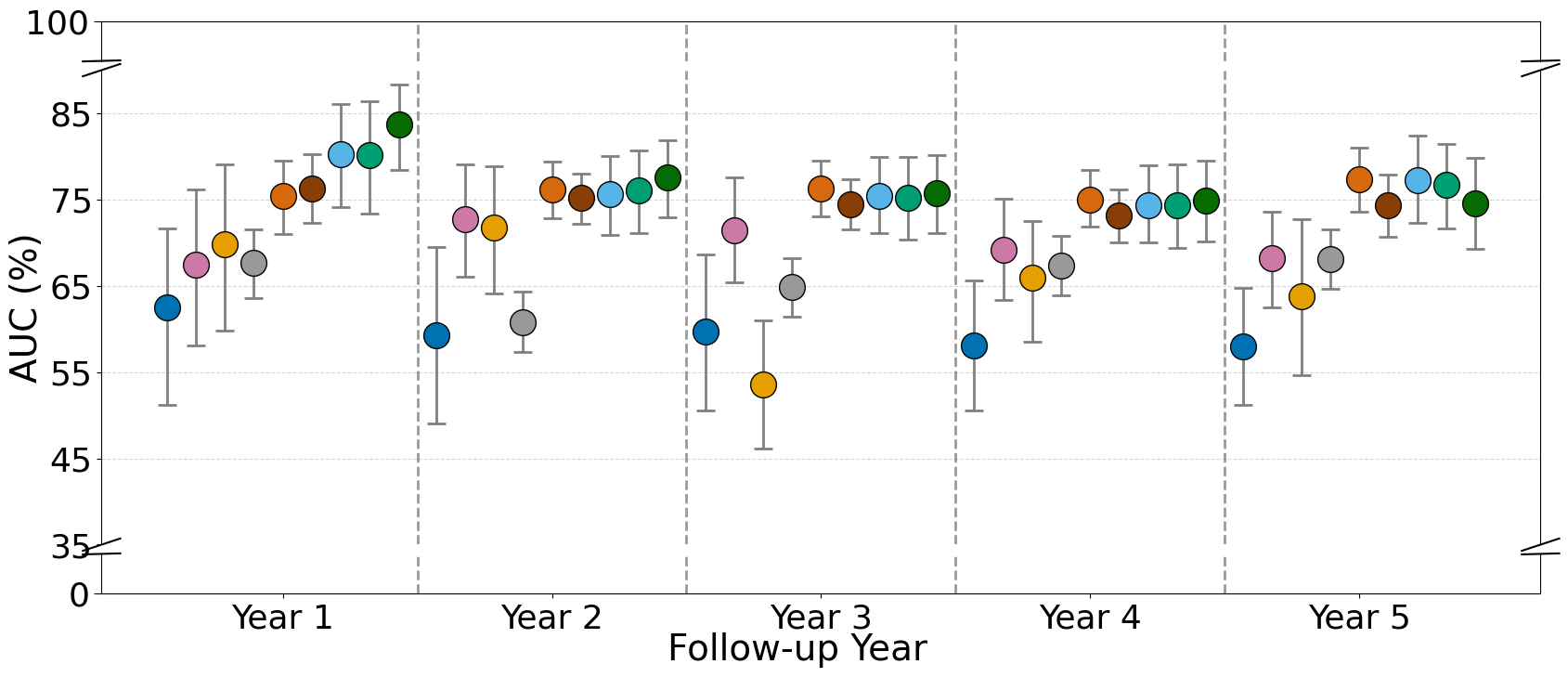}
    \caption{Medium}
\end{subfigure} \\
\begin{subfigure}[t]{0.48\textwidth}
    \centering
    \includegraphics[width=\linewidth]{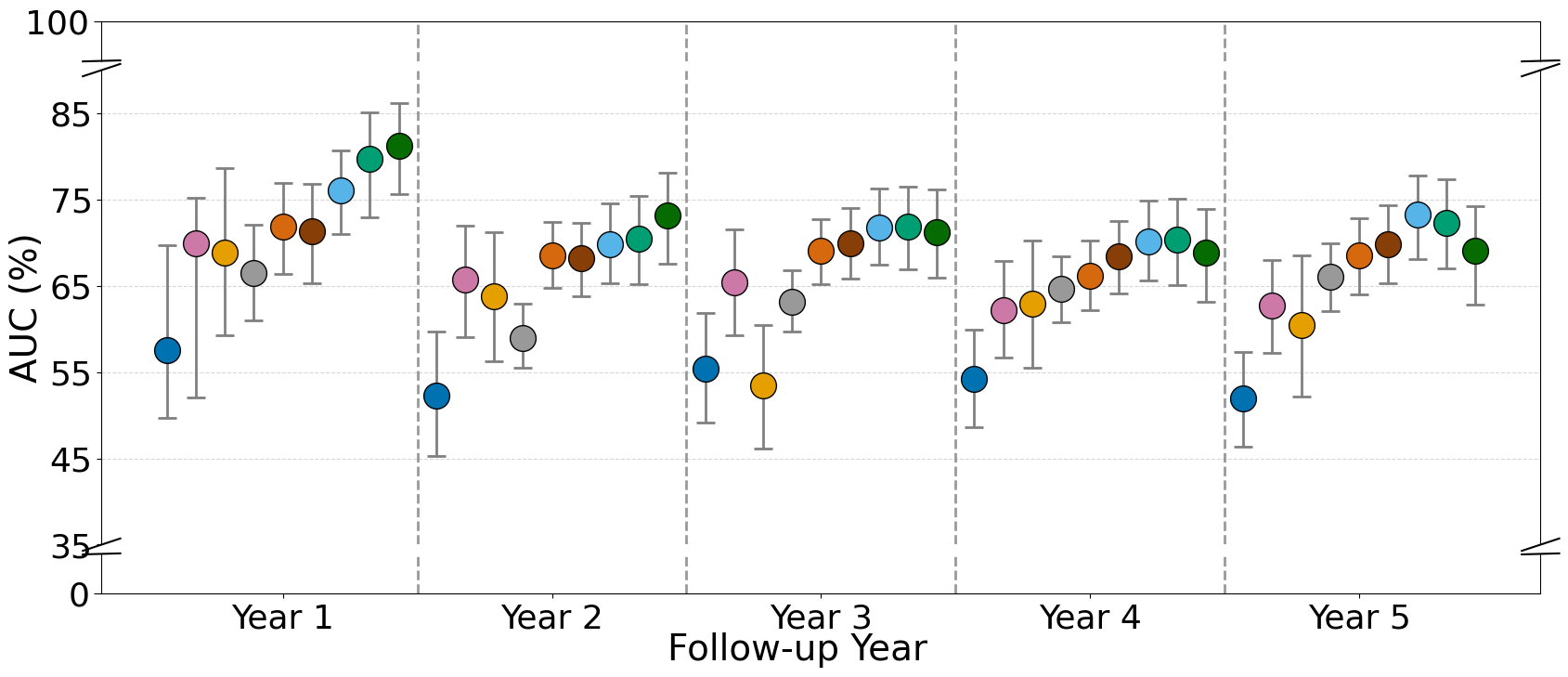}
    \caption{High}
\end{subfigure}
 \\ \vspace{0.25cm}
\includegraphics[width=0.5\linewidth]{legend.png}
\caption{Forest plots showing AUC across five follow-up years for all models stratified by breast density categories (Low, Medium, High) on the CSAW-CC dataset. Points represent mean AUC and error bars represent 95\% confidence intervals.}
\label{fig:forest_density_auc_csaw}
\end{figure}

\noindent The C-index results (Figure~\ref{fig:cindexdensity}) show strong and consistent performance of LMV-Net (\textcolor{orange}{\scalebox{0.9}{\faFire}}) and FlexLMV-Net (\textcolor{orange}{\scalebox{0.9}{\faFire}}) across breast density categories on both EMBED and CSAW-CC. On EMBED, LMV-Net performs particularly well across density groups, while FlexLMV-Net achieves comparable or higher performance in the highest-density categories. On CSAW-CC, performance differences are smaller, with both longitudinal models remaining competitive.
The year-specific AUC results (Figure~\ref{fig:forest_density_auc_embed} and Figure~\ref{fig:forest_density_auc_csaw}) show a similar trend, with both models maintaining strong performance across density categories and follow-up years. FlexLMV-Net is particularly strong for short-term prediction.
Overall, these results demonstrate robust risk stratification across density subgroups.

\subsection{Performance across cancer types}\label{performcancer}

\begin{figure}[tbph]
    \centering
    \includegraphics[width=0.65\linewidth]{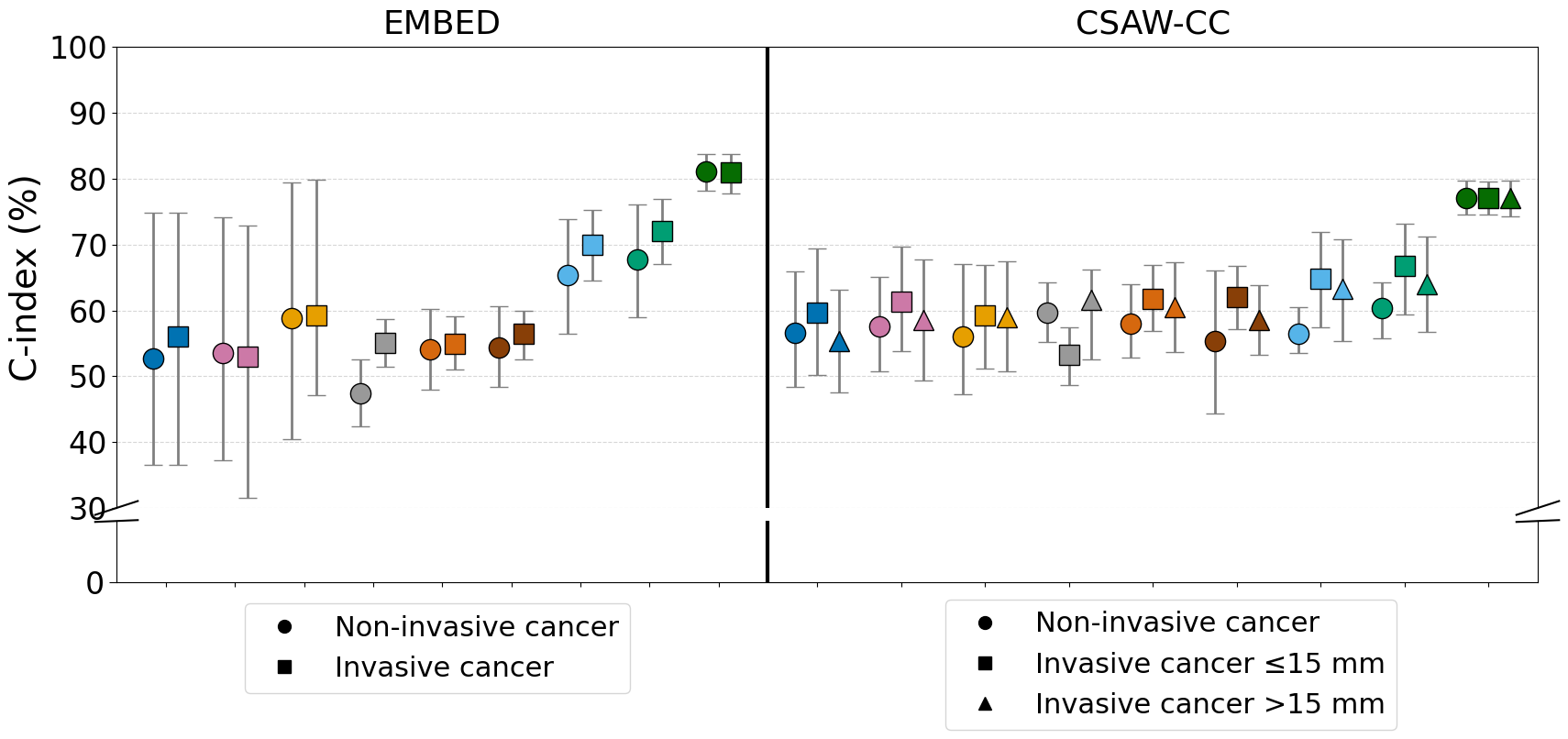}
     \\ 
\includegraphics[width=0.5\linewidth]{legend.png}
\caption{Forest plots showing  C-index for all models stratified by cancer types on the EMBED and CSAW-CC dataset. Points represent the mean C-index, and error bars indicate 95\% confidence intervals.}
\label{fig:Cancer_types}
\end{figure}
Figure~\ref{fig:Cancer_types} presents the C-index performance of LMV-Net, FlexLMV-Net, and SOTA models across cancer types on EMBED and CSAW-CC. Results are reported for non-invasive and invasive cancers, with invasive cancers in CSAW-CC further stratified by tumor size $\leq 15$ mm and $>15$ mm.
Across both datasets, FlexLMV-Net achieves the highest C-index across the cancer-type subgroups. On EMBED, it performs best for both non-invasive and invasive cancers, while on CSAW-CC it achieves the highest performance across all subgroups. Notably, the comparison between LMV-Net and FlexLMV-Net shows a substantial improvement when incorporating a larger number of prior examinations, demonstrating the benefit of FlexLMV-Net's flexible longitudinal aggregation. LMV-Net also remains competitive, generally outperforming the SOTA methods, with encoder fine-tuning providing further improvements. Overall, these results highlight the benefit of jointly modeling longitudinal and multi-view information, particularly when leveraging more extensive imaging histories, for robust risk prediction across cancer types and tumor characteristics.

\subsection{Performance across demographic subgroups}\label{subgroups}
\begin{figure}[tbp]
    \centering
    \includegraphics[width=0.6\linewidth]{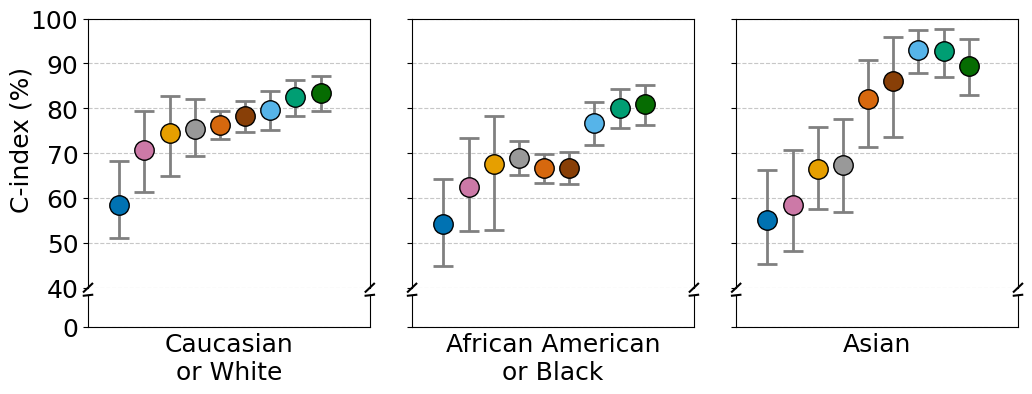}
     \\  
\includegraphics[width=0.5\linewidth]{legend.png}
    \caption{C-index values with 95\% confidence intervals for different race categories on the EMBED dataset.}
    \label{fig:emebdrace}
\end{figure}

Figure~\ref{fig:emebdrace} presents the C-index values with 95\% confidence intervals across race categories (Caucasian or White, African American or Black, and Asian) on the EMBED dataset, comparing LMV-Net, FlexLMV-Net, and SOTA models.
Across all race categories, the proposed longitudinal models generally achieve higher C-index values than the SOTA methods. FlexLMV-Net achieves the highest or near-highest performance across the evaluated groups. LMV-Net also performs competitively across all race categories, generally outperforming the SOTA methods. These results suggest that incorporating longitudinal information together with multi-view representations can improve risk prediction across diverse populations.
Performance differences between models are more pronounced for the African American or Black and Asian groups than for the Caucasian or White group.

\begin{table}[tpb]
\centering
\scriptsize
\caption{1–5 year breast cancer risk prediction performance measured by C-index and time-dependent AUC (mean $\pm$ 95\% CI). 
Models were trained on either the EMBED or CSAW-CC dataset and evaluated both on the same dataset (in-domain) and on the other dataset (cross-dataset) to assess generalization performance. Additionally, we report the cross-dataset performance drop in C-index ($\Delta$ (\%) $\downarrow$), defined as the difference between in-domain and cross-dataset performance, to quantify the degradation in predictive accuracy under domain shift.
\textcolor{blue}{\scalebox{0.9}{\faSnowflake}}: frozen encoder; 
\textcolor{orange}{\scalebox{0.9}{\faFire}}: fine-tuned encoder.}

\begin{adjustbox}{max width=\textwidth}
\begin{tabular}{lllcc*{5}{c}}
\toprule
\multicolumn{2}{c}{\textbf{Dataset}} & \textbf{Method} & \textbf{C-index (\%) $\uparrow$} &  $\Delta$ (\%) $\downarrow$ &   
\multicolumn{5}{c}{\textbf{Follow-up year AUC (\%) $\uparrow$}} \\
\cmidrule(lr){6-10}
\textbf{Training} & \textbf{Test} & & &  & \textbf{1-Y} & \textbf{2-Y} & \textbf{3-Y} & \textbf{4-Y} & \textbf{5-Y} \\
\midrule

\multirow{4}{*}{EMBED} & \multirow{4}{*}{EMBED}
& ImgFeatAlign (\textcolor{blue}{\scalebox{0.9}{\faSnowflake}})
  & \makecell{74.7\fontsize{6}{6}\selectfont$\pm$2.4}
  & 
  & \makecell{75.0\fontsize{6}{6}\selectfont$\pm$2.9} 
  & \makecell{75.5\fontsize{6}{6}\selectfont$\pm$2.4} 
  & \makecell{75.3\fontsize{6}{6}\selectfont$\pm$2.1} 
  & \makecell{75.9\fontsize{6}{6}\selectfont$\pm$2.3} 
  & \makecell{72.5\fontsize{6}{6}\selectfont$\pm$3.2} \\
& & ImgFeatAlign (\textcolor{orange}{\scalebox{0.9}{\faFire}})
&  72.2\fontsize{6}{6}\selectfont$\pm$2.7    
& 
&  72.1\fontsize{6}{6}\selectfont$\pm$3.0   
&  73.3\fontsize{6}{6}\selectfont$\pm$2.6   
&  72.2\fontsize{6}{6}\selectfont$\pm$2.5   
&  71.3\fontsize{6}{6}\selectfont$\pm$2.4  
&  72.6\fontsize{6}{6}\selectfont$\pm$3.3 \\

\cdashline{3-10} 

 &  & LMV-Net (\textcolor{blue}{\scalebox{0.9}{\faSnowflake}})
 & 78.3\fontsize{6}{6}\selectfont$\pm$3.0
 & 
 & 80.8\fontsize{6}{6}\selectfont$\pm$3.7
 & 78.3\fontsize{6}{6}\selectfont$\pm$3.5
 & 77.5\fontsize{6}{6}\selectfont$\pm$3.3
 & 76.6\fontsize{6}{6}\selectfont$\pm$3.2
 & 75.1\fontsize{6}{6}\selectfont$\pm$4.9 \\
 
 &  & LMV-Net (\textcolor{orange}{\scalebox{0.9}{\faFire}})
 & 81.4\fontsize{6}{6}\selectfont $\pm$2.8
 & 
 & 83.9\fontsize{6}{6}\selectfont $\pm$3.4
 & 82.1\fontsize{6}{6}\selectfont $\pm$3.0
 & 80.8\fontsize{6}{6}\selectfont $\pm$2.9
 & 80.0\fontsize{6}{6}\selectfont $\pm$2.8
 & 77.2\fontsize{6}{6}\selectfont $\pm$4.7
 \\

& & FlexLMV-Net (\textcolor{orange}{\scalebox{0.9}{\faFire}})  
& 81.2\fontsize{6}{6}\selectfont $\pm$2.9
& 
& 84.1\fontsize{6}{6}\selectfont $\pm$3.2
& 82.0\fontsize{6}{6}\selectfont $\pm$3.0 
& 79.6\fontsize{6}{6}\selectfont $\pm$3.1
& 79.3\fontsize{6}{6}\selectfont $\pm$3.1
& 79.0\fontsize{6}{6}\selectfont $\pm$4.4
 \\

\midrule

\multirow{4}{*}{CSAW-CC} & \multirow{4}{*}{EMBED}
& ImgFeatAlign (\textcolor{blue}{\scalebox{0.9}{\faSnowflake}})
&  64.7\fontsize{6}{6}\selectfont$\pm$2.7  
& \makecell{{\textcolor{red!70!black}{-10.0 }} }
& 66.3\fontsize{6}{6}\selectfont$\pm$3.0
& 64.8\fontsize{6}{6}\selectfont$\pm$2.8
&  65.1\fontsize{6}{6}\selectfont$\pm$2.5
&  66.3\fontsize{6}{6}\selectfont$\pm$2.5
&  64.1\fontsize{6}{6}\selectfont$\pm$4.0 \\

& & ImgFeatAlign (\textcolor{orange}{\scalebox{0.9}{\faFire}})
&   61.7\fontsize{6}{6}\selectfont$\pm$2.5    
& \makecell{{\textcolor{red!70!black}{-10.5 }} }
&   64.0\fontsize{6}{6}\selectfont$\pm$2.8
&   62.9\fontsize{6}{6}\selectfont$\pm$2.6  
&   63.2\fontsize{6}{6}\selectfont$\pm$2.5    
&   64.1\fontsize{6}{6}\selectfont$\pm$2.6   
&   65.3\fontsize{6}{6}\selectfont$\pm$3.8 \\   

\cdashline{3-10} 

 &  & LMV-Net (\textcolor{blue}{\scalebox{0.9}{\faSnowflake}})
 &  69.2\fontsize{6}{6}\selectfont$\pm$3.3
 & \makecell{{\textcolor{red!70!black}{-9.1}} }
 &  71.8\fontsize{6}{6}\selectfont$\pm$3.7
 &  69.6\fontsize{6}{6}\selectfont$\pm$3.6
 &  69.2\fontsize{6}{6}\selectfont$\pm$3.3
 &  70.5\fontsize{6}{6}\selectfont$\pm$3.4
 &  72.0\fontsize{6}{6}\selectfont$\pm$2.7 \\

 &  & LMV-Net (\textcolor{orange}{\scalebox{0.9}{\faFire}})
 & 73.7\fontsize{6}{6}\selectfont$\pm$3.3
 & \makecell{{\textcolor{red!70!black}{-7.7}} }
 & 76.7\fontsize{6}{6}\selectfont$\pm$3.9
 & 74.4\fontsize{6}{6}\selectfont$\pm$3.6
 &  73.8\fontsize{6}{6}\selectfont$\pm$3.3
  & 74.0\fontsize{6}{6}\selectfont$\pm$3.3
 &  75.5\fontsize{6}{6}\selectfont$\pm$2.8
 \\

 & & FlexLMV-Net (\textcolor{orange}{\scalebox{0.9}{\faFire}})  
& 72.9 \fontsize{6}{6}\selectfont $\pm$ 3.8
& \makecell{{\textcolor{red!70!black}{-8.3}} }
& 75.5 \fontsize{6}{6}\selectfont $\pm$ 4.6
& 73.2 \fontsize{6}{6}\selectfont $\pm$ 4.0
& 72.4 \fontsize{6}{6}\selectfont $\pm$ 3.5
& 71.1 \fontsize{6}{6}\selectfont $\pm$ 3.7
& 70.3 \fontsize{6}{6}\selectfont $\pm$ 4.6 \\
\midrule
\midrule

\multirow{4}{*}{CSAW-CC} & \multirow{4}{*}{CSAW-CC}
& ImgFeatAlign (\textcolor{blue}{\scalebox{0.9}{\faSnowflake}})
& \makecell{70.4\fontsize{6}{6}\selectfont$\pm$1.9} 
& 
& \makecell{72.0\fontsize{6}{6}\selectfont$\pm$2.4} 
& \makecell{71.5\fontsize{6}{6}\selectfont$\pm$1.9} 
& \makecell{72.6\fontsize{6}{6}\selectfont$\pm$1.9}
& \makecell{72.0\fontsize{6}{6}\selectfont$\pm$2.1} 
& \makecell{75.2\fontsize{6}{6}\selectfont$\pm$2.3} \\

& & ImgFeatAlign (\textcolor{orange}{\scalebox{0.9}{\faFire}})
&  71.9\fontsize{6}{6}\selectfont$\pm$2.0
& 
&  74.1\fontsize{6}{6}\selectfont$\pm$2.6  
&  72.3\fontsize{6}{6}\selectfont$\pm$2.0 
&  73.1\fontsize{6}{6}\selectfont$\pm$1.9   
&  72.1\fontsize{6}{6}\selectfont$\pm$2.0    
&   74.5\fontsize{6}{6}\selectfont$\pm$2.3 \\

\cdashline{3-10} 

 &  & LMV-Net (\textcolor{blue}{\scalebox{0.9}{\faSnowflake}})
 & 73.4\fontsize{6}{6}\selectfont$\pm$2.5
 & 
 & 77.5\fontsize{6}{6}\selectfont$\pm$3.3
 & 73.5\fontsize{6}{6}\selectfont$\pm$2.6
 & 74.0\fontsize{6}{6}\selectfont$\pm$2.5
 & 73.1\fontsize{6}{6}\selectfont$\pm$2.6
 & 75.9\fontsize{6}{6}\selectfont$\pm$2.8
 \\

 &  & LMV-Net (\textcolor{orange}{\scalebox{0.9}{\faFire}})
 & 74.4\fontsize{6}{6}\selectfont$\pm$2.7
 & 
 & 78.7\fontsize{6}{6}\selectfont$\pm$3.6
 & 74.1\fontsize{6}{6}\selectfont$\pm$2.9
 & 74.3\fontsize{6}{6}\selectfont$\pm$2.7
 & 73.5\fontsize{6}{6}\selectfont$\pm$2.8
 & 75.0\fontsize{6}{6}\selectfont$\pm$3.1
  \\

   & & FlexLMV-Net (\textcolor{orange}{\scalebox{0.9}{\faFire}})  
&  77.1\fontsize{6}{6}\selectfont $\pm$ 2.5
& 
& 82.5\fontsize{6}{6}\selectfont $\pm$ 3.0
& 74.6\fontsize{6}{6}\selectfont $\pm$ 3.0
& 73.7\fontsize{6}{6}\selectfont $\pm$ 2.9
& 71.9\fontsize{6}{6}\selectfont $\pm$ 3.1
& 72.3\fontsize{6}{6}\selectfont $\pm$ 3.3\\ 

\midrule

\multirow{4}{*}{EMBED} & \multirow{4}{*}{CSAW-CC}
&  ImgFeatAlign (\textcolor{blue}{\scalebox{0.9}{\faSnowflake}})
&  63.6\fontsize{6}{6}\selectfont$\pm$2.1  
& \makecell{{\textcolor{red!70!black}{-6.8}} }

&  64.9\fontsize{6}{6}\selectfont$\pm$2.5
&  64.7\fontsize{6}{6}\selectfont$\pm$2.1
&  65.2\fontsize{6}{6}\selectfont$\pm$2.0
& 64.8\fontsize{6}{6}\selectfont$\pm$2.1 
&  66.5\fontsize{6}{6}\selectfont$\pm$2.4 \\
& & ImgFeatAlign (\textcolor{orange}{\scalebox{0.9}{\faFire}})
&  64.7\fontsize{6}{6}\selectfont$\pm$2.1
& \makecell{{\textcolor{red!70!black}{-7.2}} }

&   66.8\fontsize{6}{6}\selectfont$\pm$2.7   
&   64.5\fontsize{6}{6}\selectfont$\pm$2.1    
&   65.0\fontsize{6}{6}\selectfont$\pm$2.1   
&   64.6\fontsize{6}{6}\selectfont$\pm$2.2   
&    65.3\fontsize{6}{6}\selectfont$\pm$2.5 \\

\cdashline{3-10} 

 &  &  LMV-Net (\textcolor{blue}{\scalebox{0.9}{\faSnowflake}})
 & 70.7\fontsize{6}{6}\selectfont$\pm$2.9
 & \makecell{{\textcolor{red!70!black}{-2.7}} }

 & 73.8\fontsize{6}{6}\selectfont$\pm$3.5
 & 71.0\fontsize{6}{6}\selectfont$\pm$2.9
 & 70.9\fontsize{6}{6}\selectfont$\pm$2.9
 & 70.7\fontsize{6}{6}\selectfont$\pm$3.0
 &  72.6\fontsize{6}{6}\selectfont$\pm$2.9 \\

&  & LMV-Net (\textcolor{orange}{\scalebox{0.9}{\faFire}})
 &  72.9\fontsize{6}{6}\selectfont$\pm$ 2.9
 & \makecell{{\textcolor{red!70!black}{-1.5}} }

 & 75.7\fontsize{6}{6}\selectfont$\pm$ 3.8
 & 73.1\fontsize{6}{6}\selectfont$\pm$ 3.2
 & 72.4\fontsize{6}{6}\selectfont$\pm$ 2.9
  & 71.5\fontsize{6}{6}\selectfont$\pm$ 2.9
 & 73.0\fontsize{6}{6}\selectfont$\pm$ 2.3
 \\
 & & FlexLMV-Net (\textcolor{orange}{\scalebox{0.9}{\faFire}})  
&  72.5\fontsize{6}{6}\selectfont $\pm$ 4.1
& \makecell{{\textcolor{red!70!black}{-4.6}} }
& 75.0 \fontsize{6}{6}\selectfont $\pm$ 4.0
& 72.1 \fontsize{6}{6}\selectfont $\pm$ 3.3
& 71.9 \fontsize{6}{6}\selectfont $\pm$ 3.1
& 70.9 \fontsize{6}{6}\selectfont $\pm$ 3.2
& 71.8 \fontsize{6}{6}\selectfont $\pm$ 3.1\\
\bottomrule

\end{tabular}
\end{adjustbox}
\label{tab:riskpredgeneral}
\end{table}

\subsection{Generalization performance}\label{generalperform}
\begin{figure}[tbp]
    \centering

    \begin{subfigure}[t]{0.54\linewidth}
        \centering
        \includegraphics[width=0.99\linewidth]{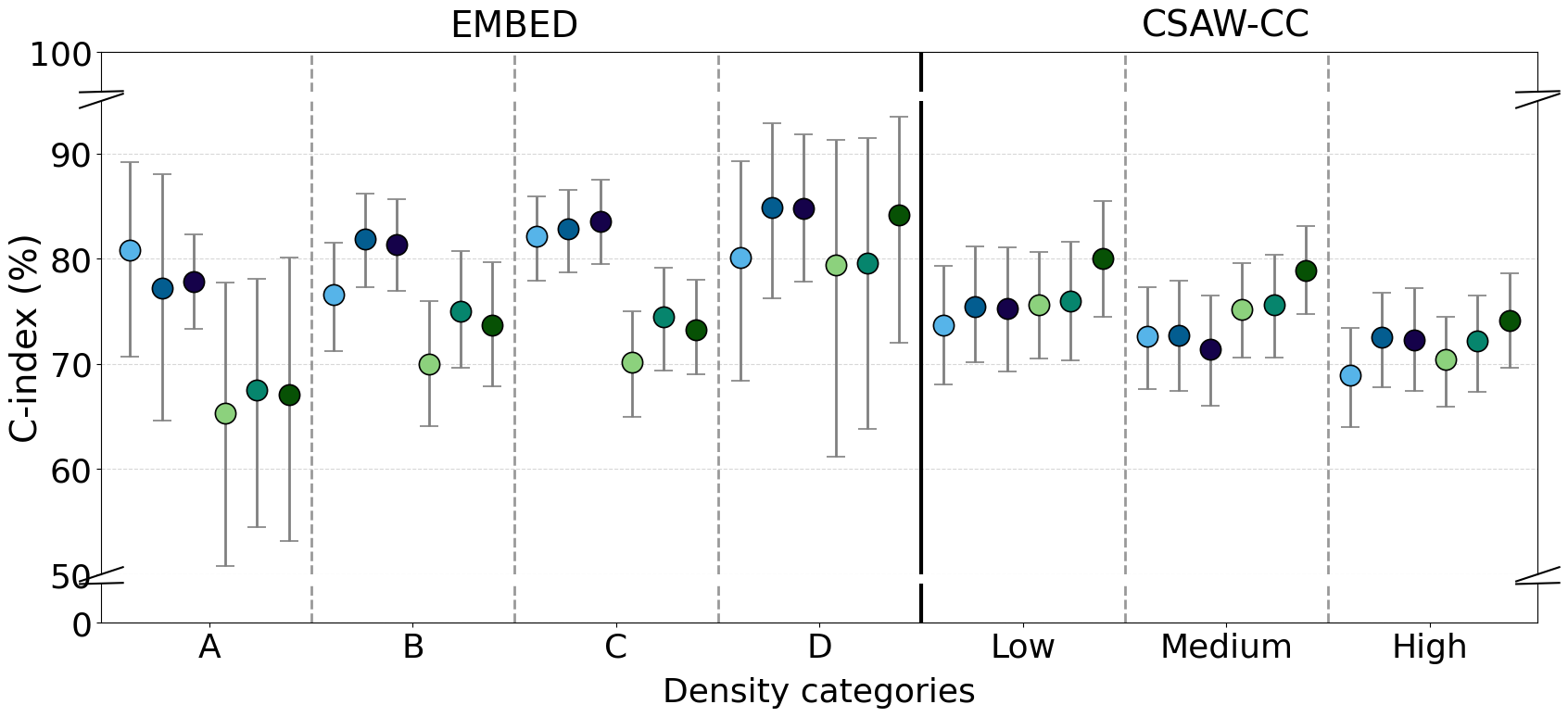}
        \caption{C-index across breast density categories.}
        \label{fig:cindexdensitygeneral}  
    \end{subfigure}
    \quad
    \begin{subfigure}[t]{0.4\linewidth}
        \centering
        \includegraphics[width=0.99\linewidth]{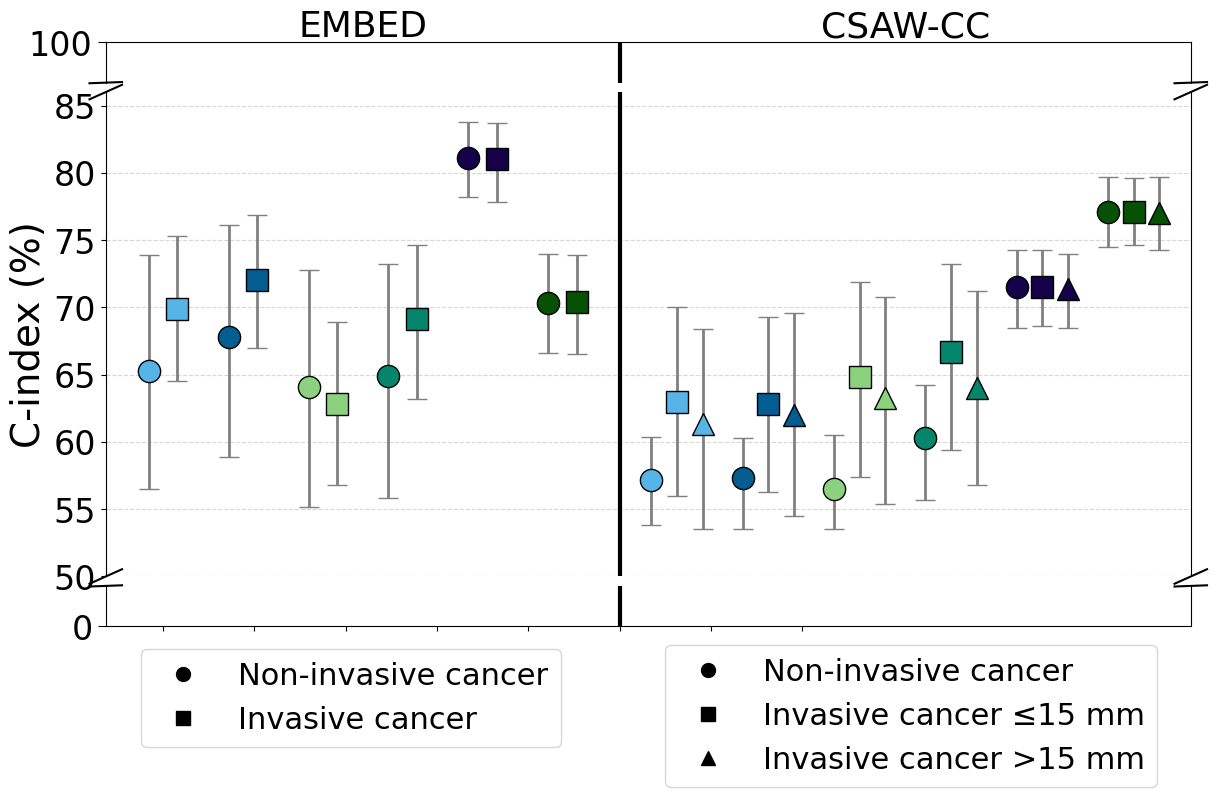}
        \caption{C-index across cancer subtypes (invasive and non-invasive).}
        \label{fig:cindex_cancer_subtypes}
    \end{subfigure}
     \\ \vspace{0.25cm}
\includegraphics[width=0.45\linewidth]{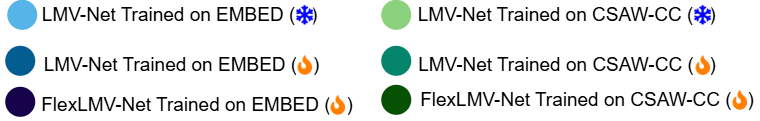}
        \caption{Generalization performance of models across (a) density categories and (b) cancer subtypes. Models are evaluated both in-domain (trained and tested on the same dataset) and cross-dataset (trained on one dataset and tested on the other) to assess generalization. Points show mean C-index and error bars indicate 95\% confidence intervals.}
      \label{fig:cindex_combined}

\end{figure}

\begin{figure}[tbp]
    \centering
    \includegraphics[width=0.5\linewidth]{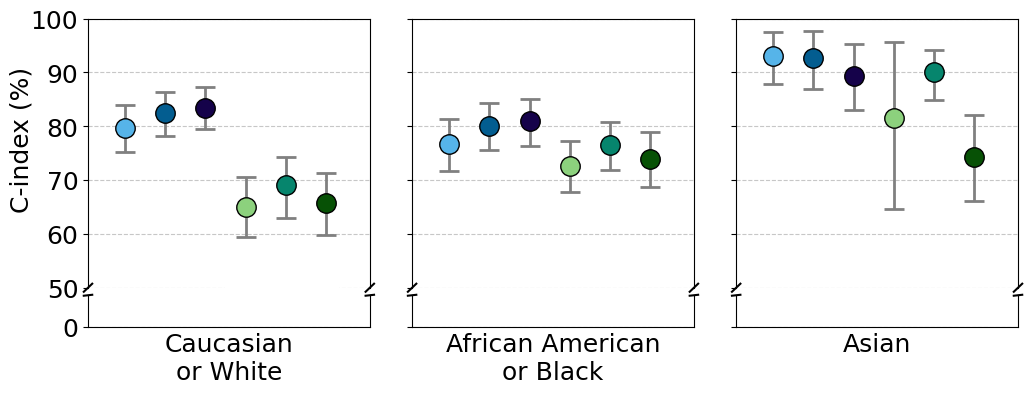}
     \\ \vspace{0.25cm}
\includegraphics[width=0.45\linewidth]{legend_gen.png}
    \caption{C-index across race categories for models evaluated on the EMBED dataset.
Models were trained either on EMBED (in-domain) or CSAW-CC (cross-dataset). Points represent the mean C-index, and error bars indicate the 95\% confidence interval.}
    \label{fig:c_index_embed_races}
\end{figure}

We evaluate the generalization performance of LMV-Net, FlexLMV-Net, and the best baseline, ImgFeatAlign. Models are trained on either EMBED or CSAW-CC and evaluated both in-domain and cross-dataset to assess robustness to domain shift. In addition to overall C-index and AUC for the 1--5 year risk prediction tasks (Table~\ref{tab:riskpredgeneral}), we evaluate C-index across breast density, cancer subtype, and race categories (Figures~\ref{fig:cindexdensitygeneral}--\ref{fig:c_index_embed_races}).
Table~\ref{tab:riskpredgeneral} shows that LMV-Net (\textcolor{orange}{\scalebox{0.9}{\faFire}}) maintains the strongest cross-dataset performance, particularly when trained on EMBED and evaluated on CSAW-CC. FlexLMV-Net achieves competitive cross-dataset performance and consistently improves in-domain performance over LMV-Net, highlighting the benefit of incorporating more prior examinations through flexible longitudinal aggregation. However, this additional history does not consistently improve cross-dataset performance, suggesting that domain shift can limit its benefits.
ImgFeatAlign generally shows a larger degradation under cross-dataset evaluation. Unlike ImgFeatAlign, which operates at the image level, LMV-Net and FlexLMV-Net model breast-level representations while explicitly integrating multi-view and longitudinal information, potentially providing greater robustness to domain shift.
Similar trends are observed across density, cancer subtype, and race subgroups. Fine-tuning generally improves cross-dataset performance, while models trained on EMBED tend to generalize more consistently to CSAW-CC than in the reverse direction. Overall, LMV-Net and FlexLMV-Net demonstrate competitive cross-dataset generalization and outperform the baseline in several settings, with the broader diversity of EMBED potentially contributing to greater robustness to the CSAW-CC domain.

\subsection{Attention map analysis}\label{attentanalysis}

\begin{figure}[hp]
    \centering
    \begin{subfigure}{0.98\linewidth}
        \includegraphics[width=\linewidth]{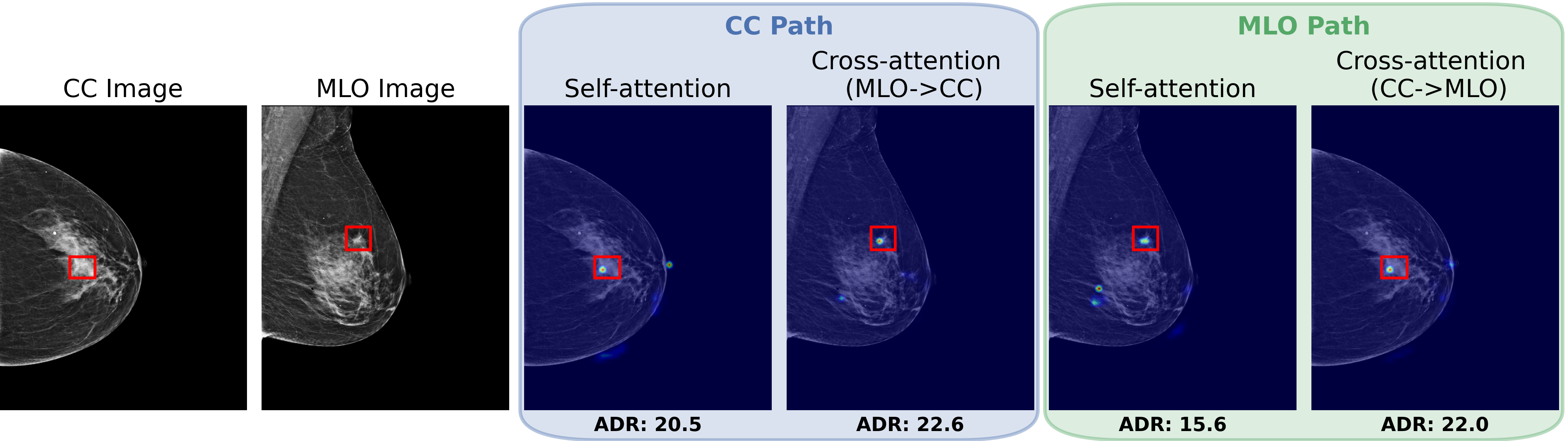}
        \caption{Patient 31110603}
    \end{subfigure}
    \hfill
    \begin{subfigure}{0.98\linewidth}
        \includegraphics[width=\linewidth]{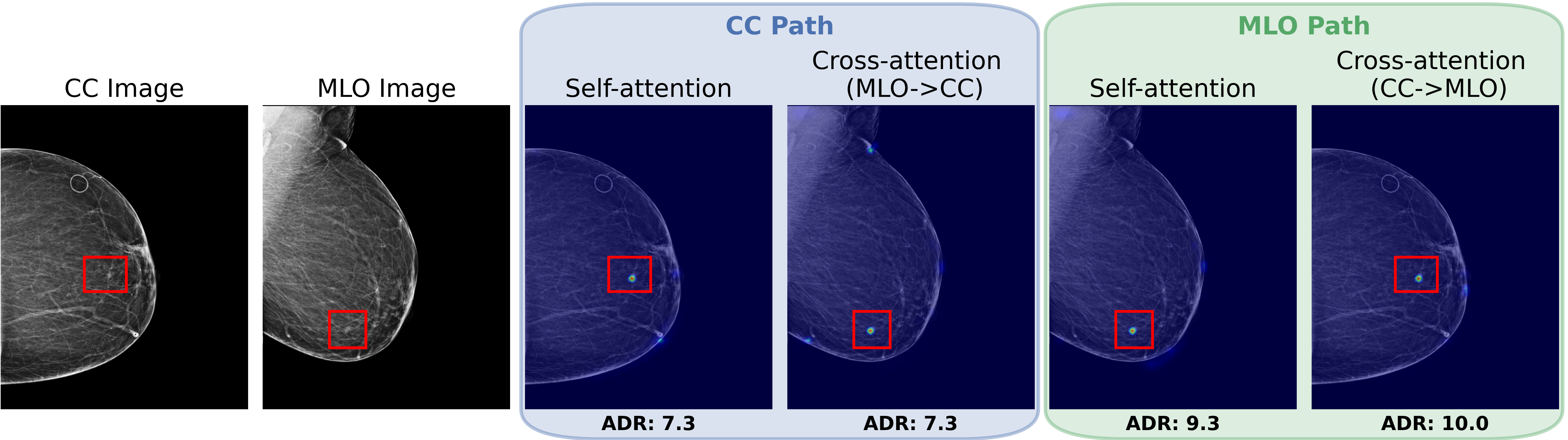}
        \caption{Patient 41292772}
    \end{subfigure}
    \hfill
    \begin{subfigure}{0.98\linewidth}
        \includegraphics[width=\linewidth]{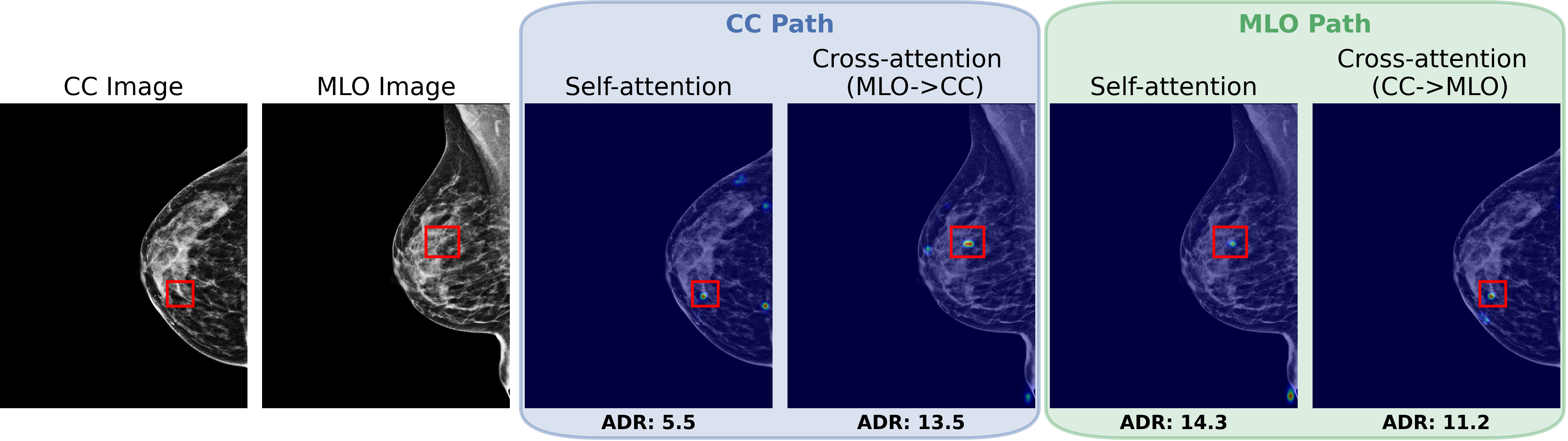}
        \caption{Patient 65755391}
    \end{subfigure}
    \hfill
    \begin{subfigure}{0.98\linewidth}
        \includegraphics[width=\linewidth]{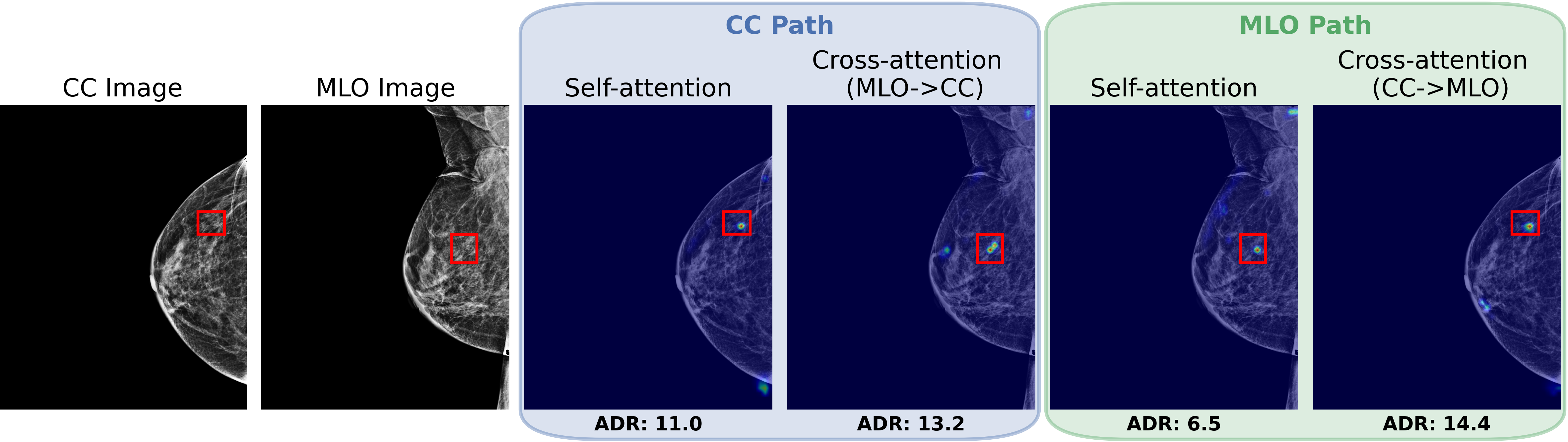}
        \caption{Patient 79200792}
    \end{subfigure}
    \caption{Attention maps across multiple cases showing that FlexLMV-Net correctly focuses on tumor regions (highlighted by red boxes).}
    \label{fig:attention}
\end{figure}

Figure~\ref{fig:attention} presents representative examples from four patients, showing mammographic images (CC and MLO views) alongside the corresponding self- and cross-attention maps produced by LMV-Net. The attention maps are overlaid on the images to highlight the regions where the proposed Dual Stream Attention block focuses. Across all cases, the model consistently assigns higher attention to suspicious areas, with particular emphasis on tumor regions (indicated by red bounding boxes provided by clinicians). Self-attention captures intra-view dependencies by concentrating on the lesion and its surrounding tissue within each view, while cross-attention leverages complementary information between CC and MLO views, attending to anatomically corresponding regions in the complementary view. Notably, this behavior is observed for tumors in both the left and right breasts (two patients each), suggesting that the model robustly identifies clinically relevant regions regardless of laterality.

We further compute the ratio of the attention density within the tumor bounding box to that within the entire breast region, referred to as the Attention Density Ratio (ADR). Inspired by the Relevance Mass Accuracy metric of~\citep{arras}, ADR quantitatively measures the concentration of model attention within the tumor region.   On average, the attention density within the tumor region is 8–12× higher than in the surrounding breast tissue. Across the four examples, the ADR is generally higher for cross-attention than for self-attention. This finding suggests that the proposed Dual Stream Attention block more effectively concentrates attention on anatomically corresponding tumor regions by leveraging information from the complementary view.

\begin{figure}[tpb]
        \centering
    \begin{subfigure}[b]{0.45\linewidth}
        \centering
        \includegraphics[width=0.85\linewidth]{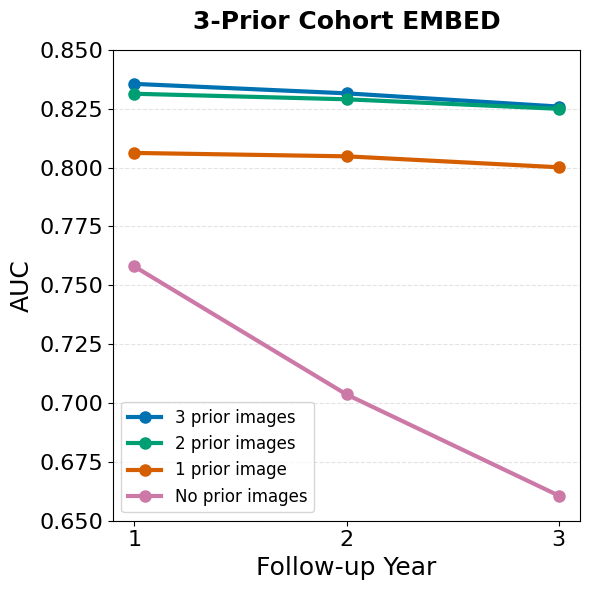}
    \end{subfigure}
    \hfill
    \begin{subfigure}[b]{0.45\linewidth}
        \centering
        \includegraphics[width=0.85\linewidth]{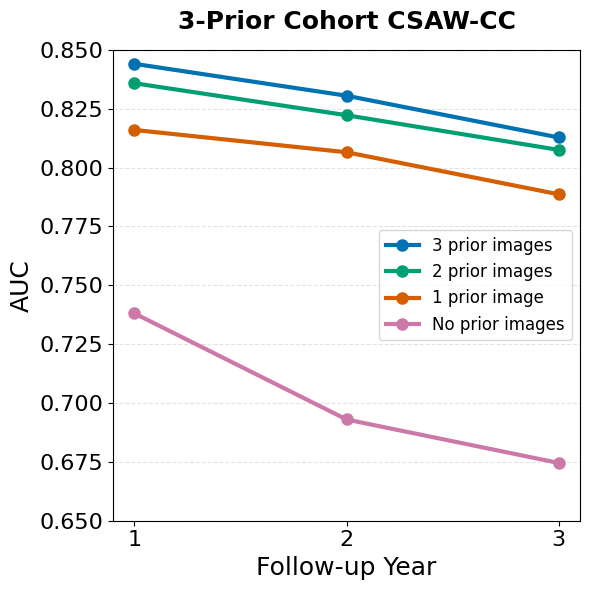}
    \end{subfigure}
    \caption{Effect of longitudinal imaging history on breast cancer risk prediction. AUC is shown across 1-, 2-, and 3-year follow-up horizons using up to three prior mammographic examinations.}
    \label{fig:longhistory}
\end{figure}

\subsection{Effect of longitudinal history length on risk prediction}\label{history}
To assess the value of longitudinal imaging history, we evaluated the model in patients with at least three prior mammographic examinations while varying the number of prior images provided as input. As shown in Figure~\ref{fig:longhistory}, incorporating longitudinal imaging history consistently improved breast cancer risk prediction compared with using no prior images. The benefit was particularly pronounced at longer follow-up horizons, where performance declined substantially when prior imaging was omitted. Performance largely plateaued when using two or three prior examinations, suggesting that even a limited longitudinal history provides substantial value for breast cancer risk prediction.

\subsection{Ablation study}\label{ablstudy}

We conducted ablation studies (Table~\ref{tab:ablation_study_final}) to assess the contribution of key components of FlexLMV-Net. The following configurations were tested: (i) \textit{Repeating current features:} temporal alignment and subtraction are removed, and the current features are concatenated three times; (ii) \textit{Concatenating with zeros:} temporal alignment and subtraction are removed and two zero tensors are concatenated with the current features; (iii) \textit{Single-view modeling:} one view branch is removed and cross-attention is replaced with self-attention; (iv) \textit{Duplicated-view modeling:} the same view is used in both branches; (v) \textit{Dual-stream attention removed:} the dual-stream attention block is removed; (vi) \textit{Multi-view learning:} only $Y_{\text{multi-view}}$ is used; (vii) \textit{Encoder fine-tuning:} the pretrained Mirai backbone is frozen; and (viii) \textit{Single prior:} only one prior examination is used, corresponding to LMV-Net where we don't have the feature-aware temporal fusion block; and
(ix) \textit{Multiple priors:} the full FlexLMV-Net configuration uses up to three prior examinations.

The results demonstrate that removing longitudinal information, multi-view modeling, dual-stream attention, or encoder fine-tuning consistently reduces performance. Using a single prior also performs worse than the full FlexLMV-Net, highlighting the benefit of incorporating multiple prior examinations. Overall, the full model achieves the best performance on both EMBED and CSAW-CC, demonstrating the complementary contribution of its key components.

\begin{table*}[t]
\centering
\scriptsize
\caption{Ablation study for 1–5 year risk: mean C-index and AUC on EMBED and CSAW-CC. MP: Using multiple prior images (up to 3), MV: Multi-view, DV: Different views in both branches, FT: Encoder fine-tuning, DSA: Dual Stream Attention.}
\begin{adjustbox}{max width=\textwidth}
\begin{tabular}{llccccccccccccc}
\toprule
\multirow{2}{*}{} &
\multirow{2}{*}{\textbf{Variant}} &
\multirow{2}{*}{\textbf{Prior}} &
\multirow{2}{*}{\textbf{MP}} &
\multirow{2}{*}{\textbf{MV}} &
\multirow{2}{*}{\textbf{DV}} &
\multirow{2}{*}{\textbf{FT}} &
\multirow{2}{*}{\textbf{DSA}} &
\multirow{2}{*}{\makecell{\textbf{Y$_{\text{MLO}}$}\\ \textbf{Y$_{\text{CC}}$}}} &
\multirow{2}{*}{\textbf{C-index (\%) $\uparrow$}} &
\multicolumn{5}{c}{\textbf{Follow-up year AUC (\%) $\uparrow$}} \\

\cmidrule{11-15}
&&&&&&&&&& \textbf{1-Y} & \textbf{2-Y} & \textbf{3-Y} & \textbf{4-Y} & \textbf{5-Y} \\ 
\midrule
\multirow{8}{*}{\rotatebox{90}{\textbf{EMBED}}}
& (i) No prior (current repeated)        & \xmark & \xmark & \cmark & \cmark & \cmark & \cmark & \cmark & 77.8 & 80.3 & 78.1 & 76.9 & 76.3 & 73.6 \\
& (ii) No prior (zero-padded)   & \xmark &\xmark &  \cmark & \cmark & \cmark & \cmark & \cmark & 77.5  & 80.2 & 78.3 & 76.4 & 76.2 & 73.8 \\
& (iii) Only one view          & \cmark & \xmark & \xmark & \cmark & \cmark & \cmark & \cmark & 70.8 & 72.2 & 70.8 & 68.9 & 65.9 & 61.9 \\
& (iv) Same view both branches& \cmark & \xmark & \cmark & \xmark & \cmark & \cmark & \cmark & 77.4  & 78.8 & 78.1 & 77.7 & 76.2 & 74.7 \\
& (v) Without DSA     & \cmark & \xmark & \cmark & \cmark & \cmark & \xmark & \cmark & 72.8 & 73.9 & 73.7 & 72.8 & 70.3 & 65.8 \\
& (vi) Use only Y$_\text{multi-view}$ & \cmark & \xmark  & \cmark & \cmark & \cmark & \cmark & \xmark & 80.1 & 83.0 & 80.4 & 79.4 & 78.8 & 76.5 \\
& (vii) No finetuning           & \cmark & \xmark &  \cmark & \cmark & \xmark & \cmark & \cmark & 78.3 & 80.8 & 78.3 & 77.5 & 76.6 & 75.1 \\
& (viii) One Prior & \cmark &\xmark &  \cmark & \cmark & \cmark & \cmark & \cmark & \textbf{81.4} & \textbf{83.9} & \textbf{82.1} & \textbf{80.8} & \textbf{80.0} & 77.2 \\
& (ix) Multiple Prior & \cmark &\cmark &  \cmark & \cmark & \cmark & \cmark & \cmark & \textbf{81.2} & \textbf{84.1} & \textbf{82.0} & \textbf{79.6} & \textbf{79.3} & \textbf{ 79.0} \\ 
\midrule
\multirow{8}{*}{\rotatebox{90}{\textbf{CSAW-CC}}} 
& (i) No prior (current repeated)        & \xmark & \xmark & \cmark & \cmark & \cmark & \cmark & \cmark & 73.0  & 75.3 & 73.8 & 73.2 & 72.3 & 73.2 \\
& (ii) No prior (zero-padded)   & \xmark & \xmark &  \cmark & \cmark & \cmark & \cmark & \cmark & 72.2  & 75.2 & 73.0 & 72.6 & 71.9 & 72.4 \\
& (iii) Only one view          & \cmark & \xmark & \xmark & \cmark & \cmark & \cmark & \cmark & 73.3  & 73.8 & 73.1 & 73.5 & 73.1 & 73.9 \\
& (iv) Same view both branches& \cmark & \xmark & \cmark & \xmark & \cmark & \cmark & \cmark & 72.2  & 75.2 & 72.6 & 72.5 & 71.1 & 73.0 \\
& (v) Without DSA     & \cmark & \xmark & \cmark & \cmark & \cmark & \xmark & \cmark & 69.5  & 70.5 & 70.0 & 70.6 & 69.3 & 71.4 \\
& (vi) Use only Y$_\text{multi-view}$ & \cmark & \xmark & \cmark & \cmark & \cmark & \cmark & \xmark & 73.3 & 77.6 & 73.3 & 73.4 & 72.6 & 74.4 \\
& (vii) No finetuning           & \cmark & \xmark & \cmark & \cmark & \xmark & \cmark & \cmark & 73.4  & 77.5 & 73.5 & 74.0 & 73.1 & \textbf{75.9} \\
& (viii) One Prior & \cmark & \xmark & \cmark & \cmark & \cmark & \cmark & \cmark & 74.4 & 78.7 & \textbf{74.1} & \textbf{74.3} & \textbf{73.5} & 75.0 \\
& (ix) Multiple Prior & \cmark & \cmark &  \cmark & \cmark & \cmark & \cmark & \cmark & \textbf{77.1} & \textbf{82.5} & \textbf{74.6} & \textbf{73.7} & 71.9 & 72.3 \\ 
\bottomrule
\end{tabular}
\end{adjustbox}
\label{tab:ablation_study_final}
\end{table*}

\section{Limitations and future works}\label{limitationsoutlook}
While LMV-Net and FlexLMV-Net demonstrate promising results, several limitations remain. First, further validation on independent multi-center and prospective cohorts is needed to assess generalizability across populations, imaging protocols, and clinical settings. Second, the explicit feature-space alignment depends on accurate spatial correspondence, and alignment errors may propagate to risk prediction. Future work could explore more robust, anatomy-aware, or jointly learned alignment strategies. Third, although FlexLMV-Net supports a variable number of prior examinations, future work could investigate longer screening histories and strategies for optimally selecting and weighting prior examinations. Incorporating bilateral breast information is another promising direction, as between-breast asymmetries may provide complementary risk-related signals. Finally, extending the framework to digital breast tomosynthesis (DBT) could leverage its three-dimensional structural information and enable investigation of its complementary value alongside conventional mammography.

\section{Conclusion}\label{conclusion}
In this study, we introduced FlexLMV-Net, a longitudinal multi-view framework that supports a variable number of prior examinations through explicit longitudinal feature alignment and feature-aware temporal fusion of CC and MLO views for breast cancer risk prediction. LMV-Net can be viewed as a special case of FlexLMV-Net with $K=1$ prior examination, providing a fixed-length formulation of the framework. Both frameworks demonstrated strong performance across the EMBED and CSAW-CC datasets and diverse subgroups, while additional longitudinal history further improved risk prediction. In addition, we introduced BCRBench, an open-source benchmark repository that unifies breast cancer risk prediction models within a standardized framework to facilitate fair and reproducible comparisons. Together, these contributions demonstrate the potential of longitudinal multi-view modeling for breast cancer risk stratification and provide a foundation for future research toward more personalized screening and earlier detection.

\section*{CRediT authorship contribution statement}
\textbf{Solveig Thrun:} Conceptualization, Software, Methodology, Formal analysis, Investigation, Visualization, Writing - original draft, Writing - review \& editing.  \textbf{ Zijun Sun:} Methodology, Writing - review \& editing. \textbf{Suaiba A. Salahuddin:} Methodology, Writing - review \& editing. \textbf{Kristoffer Wickstrøm:}  Conceptualization, Methodology, Writing - review \& editing, Supervision.  \textbf{Elisabeth Wetzer:}  Conceptualization, Methodology, Writing - review \& editing, Supervision. \textbf{Stine Hansen:} Conceptualization, Methodology, Writing - review \& editing, Supervision. \textbf{Robert Jenssen:}  Conceptualization, Methodology, Writing - review \& editing, Supervision, Funding acquisition.  \textbf{Michael Kampffmeyer:} Conceptualization, Methodology, Writing - review \& editing, Supervision,  Project administration, Funding acquisition.

\section*{Declaration of Competing Interest}
The authors declare that they have no known competing financial interests or personal relationships that could have appeared to influence the work reported in this paper.

 \section*{Data availability}
The source code of FlexLMV-Net is publicly available at  \url{https://github.com/sot176/FlexLMV-Net} and the proposed BCRBench benchmarking repository, which includes SOTA breast cancer risk prediction models, is available at\url{https://github.com/sot176/BCRBench}. All datasets used in this work are publicly accessible and links to request access are provided in the Experimental Setup section.
 
\section*{Acknowledgements}
This work was supported by The Research Council of Norway (Visual Intelligence, grant no. 309439 as well as FRIPRO grant no. 315029 and 360068 and IKTPLUSS grant no. 303514). We acknowledge Sigma2 – the National Infrastructure for High Performance Computing and Data Storage in Norway, for awarding this project access to the LUMI supercomputer, owned by the EuroHPC Joint Undertaking, hosted by CSC (Finland) and the LUMI consortium through Sigma2 [grant no. NN8106K]. We would like to thank the Research Software Engineering (RSE) Group at UiT The Arctic University of Norway for their valuable support and advice in designing and implementing the unified GitHub repository BCRBench.

\appendix

\section{Dataset Details and Dataset Splits}
Tables~\ref{tab:timecancerdataset} and~\ref{tab:combined_densitydistr} summarize the Time-to-Cancer label distribution, dataset split, and breast density distribution for the EMBED and CSAW-CC datasets.
For EMBED, breast density labels are unavailable for some cases. Therefore, the number of cases reported in the density distribution table does not match the total number of cases reported in the Time-to-Cancer table.

\begin{table}[H]
\centering
\scriptsize
\caption{Time-to-Events label distribution and dataset splits for breast-level CC+MLO exam pairs. Percentages are reported relative to the total number of cases per split.}
\label{tab:timecancerdataset}
\begin{adjustbox}{center}
\begin{tabular}{llccccccccc}
\toprule
& & \multicolumn{6}{c}{\textbf{Positive: Time-to-Events}} & \textbf{Negative} & \\
\cmidrule(lr){3-8} \cmidrule(lr){9-9}
\textbf{Dataset} & \textbf{Split} & \textbf{1Y} & \textbf{2Y} & \textbf{3Y} & \textbf{4Y} & \textbf{5Y} & \textbf{All} & \textbf{$>$5Y} & \textbf{Total} \\
\midrule

\multirow{6}{*}{\textbf{EMBED}} 
& Train &
\makecell{370\\(2.9\%)} &
\makecell{96\\(0.8\%)} &
\makecell{83\\(0.6\%)} &
\makecell{46\\(0.4\%)} &
\makecell{34\\(0.3\%)} &
\makecell{629\\(4.9\%)} &
\makecell{12262\\(95.1\%)} &
\makecell{12891}\\

& Val &
\makecell{146\\(2.9\%)} &
\makecell{32\\(0.6\%)} &
\makecell{29\\(0.6\%)} &
\makecell{20\\(0.4\%)} &
\makecell{14\\(0.3\%)} &
\makecell{241\\(4.7\%)} &
\makecell{4907\\(95.3\%)} &
\makecell{5148}\\

& Test &
\makecell{220\\(2.9\%)} &
\makecell{76\\(1.0\%)} &
\makecell{54\\(0.7\%)} &
\makecell{37\\(0.5\%)} &
\makecell{29\\(0.4\%)} &
\makecell{416\\(5.4\%)} &
\makecell{7341\\(94.6\%)} &
\makecell{7757}\\

& \textbf{Total} &
\makecell{\textbf{736}\\\textbf{(2.9\%)}} &
\makecell{\textbf{204}\\\textbf{(0.8\%)}} &
\makecell{\textbf{166}\\\textbf{(0.7\%)}} &
\makecell{\textbf{103}\\\textbf{(0.4\%)}} &
\makecell{\textbf{77}\\\textbf{(0.3\%)}} &
\makecell{\textbf{1286}\\\textbf{(5.0\%)}} &
\makecell{\textbf{24510}\\\textbf{(95.0\%)}} &
\makecell{\textbf{25796}} \\

\midrule

\multirow{6}{*}{\textbf{CSAW-CC}} 
& Train &
\makecell{352\\(2.1\%)} &
\makecell{350\\(2.1\%)} &
\makecell{94\\(0.6\%)} &
\makecell{224\\(1.3\%)} &
\makecell{88\\(0.5\%)} &
\makecell{1108\\(6.6\%)} &
\makecell{15686\\(93.4\%)} &
\makecell{16794}\\

& Val &
\makecell{140\\(2.1\%)} &
\makecell{142\\(2.1\%)} &
\makecell{48\\(0.7\%)} &
\makecell{92\\(1.4\%)} &
\makecell{28\\(0.4\%)} &
\makecell{450\\(6.8\%)} &
\makecell{6188\\(93.2\%)} &
\makecell{6638}\\

& Test &
\makecell{208\\(2.1\%)} &
\makecell{214\\(2.1\%)} &
\makecell{76\\(0.8\%)} &
\makecell{124\\(1.2\%)} &
\makecell{54\\(0.5\%)} &
\makecell{676\\(6.7\%)} &
\makecell{9402\\(93.3\%)} &
\makecell{10078}\\

& \textbf{Total} &
\makecell{\textbf{700}\\\textbf{(2.1\%)}} &
\makecell{\textbf{706}\\\textbf{(2.1\%)}} &
\makecell{\textbf{218}\\\textbf{(0.7\%)}} &
\makecell{\textbf{440}\\\textbf{(1.3\%)}} &
\makecell{\textbf{170}\\\textbf{(0.5\%)}} &
\makecell{\textbf{2234}\\\textbf{(6.6\%)}} &
\makecell{\textbf{31276}\\\textbf{(93.4\%)}} &
\makecell{\textbf{33510}} \\

\bottomrule
\end{tabular}
\end{adjustbox}
\end{table}

\begin{table}[H]
\centering
\scriptsize
\caption{Breast density distribution across the EMBED and CSAW-CC datasets for breast-level CC+MLO exam pairs. Percentages are reported relative to the total number of cases per split.}
\label{tab:combined_densitydistr}
\begin{adjustbox}{center}
\begin{tabular}{lccccccc}
\toprule
& \multicolumn{4}{c}{\textbf{EMBED (BI-RADS)}} & \multicolumn{3}{c}{\textbf{CSAW-CC}} \\
\cmidrule(lr){2-5} \cmidrule(lr){6-8}
\textbf{Split} & \textbf{A} & \textbf{B} & \textbf{C} & \textbf{D} & \textbf{Low} & \textbf{Medium} & \textbf{High} \\
\midrule

Train &
\makecell{1435\\(11.1\%)} &
\makecell{5631\\(43.7\%)} &
\makecell{5214\\(40.4\%)} &
\makecell{569\\(4.4\%)} &
\makecell{5776\\(34.4\%)} &
\makecell{5538\\(33.0\%)} &
\makecell{5480\\(32.6\%)} \\

Val &
\makecell{587\\(11.4\%)} &
\makecell{2164\\(42.0\%)} &
\makecell{2119\\(41.2\%)} &
\makecell{265\\(5.1\%)} &
\makecell{2130\\(32.1\%)} &
\makecell{2266\\(34.1\%)} &
\makecell{2242\\(33.8\%)} \\

Test &
\makecell{943\\(12.2\%)} &
\makecell{3279\\(42.3\%)} &
\makecell{3174\\(40.9\%)} &
\makecell{344\\(4.4\%)} &
\makecell{3354\\(33.3\%)} &
\makecell{3480\\(34.5\%)} &
\makecell{3244\\(32.2\%)} \\

\midrule

\textbf{Total} &
\makecell{\textbf{2965}} &
\makecell{\textbf{11074}} &
\makecell{\textbf{10507}} &
\makecell{\textbf{1178}} &
\makecell{\textbf{11260}} &
\makecell{\textbf{11284}} &
\makecell{\textbf{10966}} \\

\bottomrule
\end{tabular}
\end{adjustbox}
\end{table}




\end{document}